\RequirePackage{fix-cm}
\documentclass[twocolumn]{svjour3}          
\smartqed  
\usepackage{url}
\usepackage{graphicx}        

\usepackage{amsmath,bm,amsfonts,amssymb}

\usepackage{commath}      
\usepackage{color,xcolor,ucs}
\usepackage{subfig}
\usepackage[font=footnotesize,labelfont=bf]{caption}
\usepackage{floatrow}
\usepackage{tabularx}
\usepackage{float}
\usepackage{cite}
\usepackage{xr-hyper}
\usepackage{hyperref}

\usepackage{wrapfig}

\usepackage{algorithm}
\usepackage{algpseudocode}
\usepackage{multirow}
\usepackage{lipsum}
\usepackage{textcomp} 
\newcommand{\B}{\textbf}

\definecolor{revisioncolor}{rgb}{0.1,0.1,1}

\begin{document}
\sloppy 
\title{An Integrated Localisation, Motion Planning and Obstacle Avoidance Algorithm in Belief Space
}


\author{Antony Thomas \and Fulvio Mastrogiovanni
\and Marco Baglietto}


\institute{The authors are with the Department of Informatics, Bioengineering, Robotics, and Systems Engineering, University of Genoa, Via All'Opera Pia 13, 16145 Genoa, Italy. \\
             \email{antony.thomas@dibris.unige.it}}           

\date{Received: date / Accepted: date}

\maketitle

\begin{abstract}
As robots are being increasingly used in close proximity to humans and objects, it is imperative that robots operate safely and efficiently under real-world conditions. Yet, the environment is seldom known perfectly. Noisy sensors and actuation errors compound to the errors introduced while estimating features of the environment. We present a novel approach (1) to incorporate these uncertainties for robot state estimation and  (2) to compute the probability of collision pertaining to the estimated robot configurations. The expression for collision probability is obtained as an infinite series and we prove its convergence. An upper bound for the truncation error is also derived and the number of terms required is demonstrated by analyzing the convergence for different robot and obstacle configurations. We evaluate our approach using two simulation domains which use a roadmap-based strategy to synthesize trajectories that satisfy collision probability bounds. 
\end{abstract}

\section{Introduction}
Planning and decision making under uncertainty are fundamental requirements for autonomous robots. Uncertainties often arise due to insufficient knowledge about the environment, imperfect sensing and inexact robot motion. In these conditions, the robot poses or other variables of interest can only be dealt with in terms of probabilities. Planning is therefore performed in the \textit{belief} space, which corresponds to the set of all probability distributions over possible robot states~\cite{platt2010RSS}. At a given time instant, we consider the belief or the belief state of the robot which corresponds to a probability distribution of the robot state (or other variables of interest) given the measurements and controls thus far~\cite{van_den_berg2012IJRR}. Consequently, for efficient planning and decision making, it is required to reason about future belief distributions due to candidate actions and the corresponding expected observations. Such a problem falls under the category of \textit{Partially Observable Markov Decision Processes} (POMDPs)~\cite{kaelbling1998AI}. 

\begin{figure}[t]
	\centering
		\includegraphics[scale=0.15]{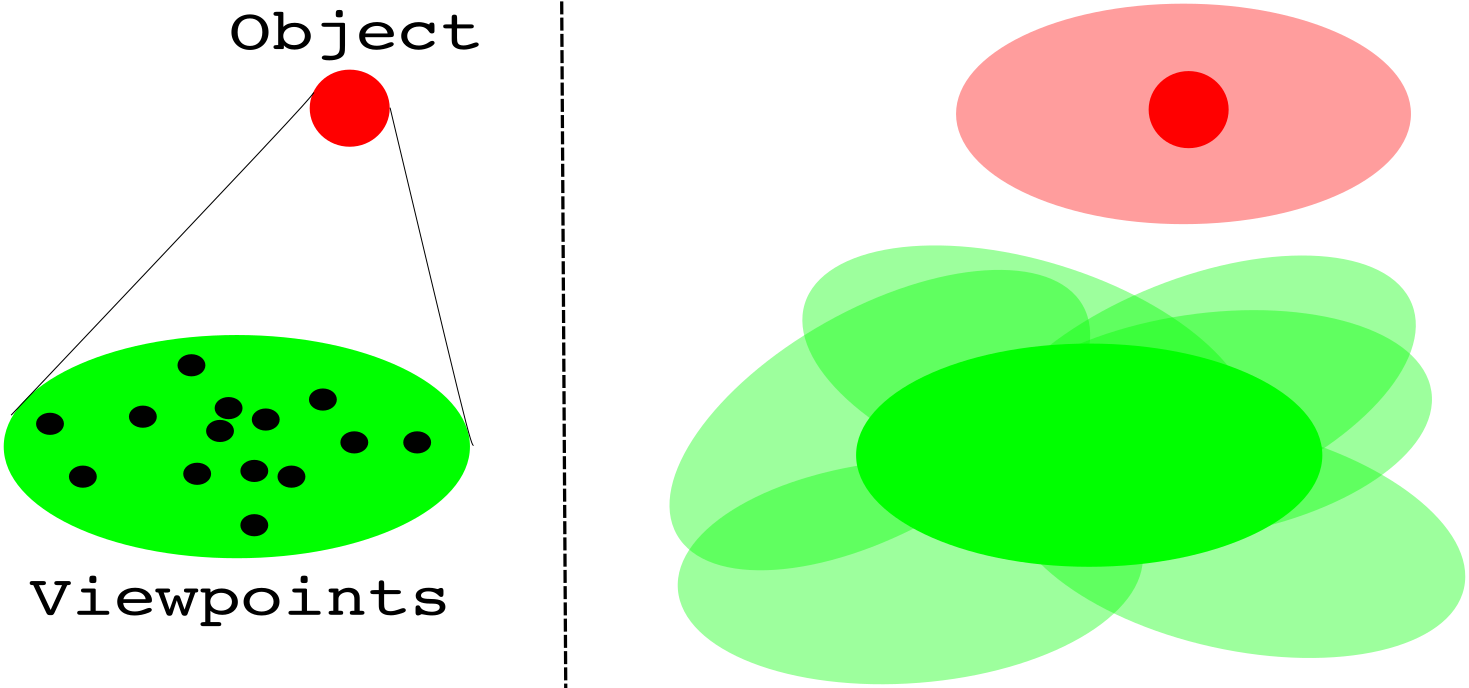}
		\caption{The red blob denotes an object in the environment. The green region corresponds to the set of viewpoints from which the object can be observed; some of these viewpoints are shown as black dots. On the right hand side, the red shaded region denotes the uncertainty in object location, with the red blob denoting its mean position. The corresponding viewpoint region is visualized as the intersection between different viewpoint regions that correspond to the object being at different locations (left hand side shows one instance of this).}
	\label{fig:concept}
\end{figure}

Robots are becoming ubiquitous in our day-to-day lives and are being increasingly used in close proximity to humans and other objects in service-oriented scenarios such as factories, living spaces, or elderly care facilities. It is therefore of vital importance that robots operate efficiently and safely in real-world conditions. Localization is a key aspect for safe and efficient robot motion as it is a precursor to solving the problems ``where to move to" and ``how to reach there". A robot perceives the environment through its sensors and distinct objects known as landmarks aid the robot in localizing. However, most approaches assume that these landmarks are known with high certainty. For example, given the map of the environment, while planning for future actions the standard Markov localization\footnote{The application of Bayes filter to the localization problem is called Markov localization~\cite{thrun2005book}.} does not take into account the map uncertainty, that is, the landmark location uncertainties are ignored and the locations are assumed to be perfectly known. This means that given the map and the sensor range\footnote{Note that the concepts discussed here are applicable to any sensor used for robot localization. In particular, in this work (Section~\ref{results}) we use a laser range finder and beacons that give signal measurements in terms of the distance to the beacons.}, for any landmark, there exists a set of viewpoints from which an observation may be obtained. Let us consider for example a robot equipped with a laser range finder and observing a landmark. Whenever the robot location is such that the landmark falls within the sensing range, a measurement is obtained. Thus, there exists a set of robot locations or viewpoints from which a measurement of the landmark may be obtained. Therefore, when the landmark locations are assumed to be perfect, this set of viewpoints can be easily determined since it depends on the environment map and the sensing capabilities of the sensor employed. Yet, this might not be true in practice. For example, consider the map of an environment obtained from a \textit{Simultaneous Localization and Mapping} (SLAM) session. Due to the dynamic nature of the environment, the objects of interests could be occluded when viewed from the set of viewpoints which would have otherwise produced a full observation. Moreover, an erroneous localization, for example due to wrong data association, could lead to wrongly estimated object poses. Thus, in such cases it is more fitting to consider the uncertainty in the landmark locations. This landmark uncertainty directly translates to the fact that the viewpoints whence the object can be observed is uncertain. This is visualized in Fig.~\ref{fig:concept}. As seen on the left hand side of the figure, when the object location is known perfectly, there exists a region (green) from which the object can be observed. Note that as discussed before this region is determined from the environment map and the sensor capabilities. Some of the viewpoints inside this region are shown in black. On the right hand side of the figure, we consider the uncertainty in landmark location the red shaded region denotes the uncertainty in landmark location. Since the object location is not known precisely (object can be anywhere within the uncertainty region), given a viewpoint, it cannot be said with certainty that the object will be observed. This is so because given a viewpoint, a landmark is observed if it falls within the sensing range. However, since the landmark location is not fixed and is uncertain, the landmark may or may not be within the sensing range. For example, if the landmark location is Gaussian distributed, then the landmark, in practice can be anywhere within the (say) 3-$\sigma$ uncertainty region. Thus we cannot define a precise region from which the landmark can be observed. Therefore, one can only reason in terms of the probability of observing the object from the considered viewpoint. This results in a probability distribution function for the viewpoints. Consequently, not considering this uncertainty can wrongly localize the robot, leading to inefficient plans causing catastrophes. From now on, we will use the term \textit{object uncertainty} to refer to the notion of uncertainty in landmark location. 

In order to ensure safe robots' motion, it is also essential to consider collision avoidance strategies. As robots are being increasingly used in service-oriented scenarios with both static and dynamic obstacles, deterministic approaches do not fare well. Moreover, in the case of dynamic obstacles, their future states have to be predicted. Yet this is an added difficulty due to the lack of perfect knowledge of their motions. As a result, providing safety guarantees is difficult. 

\subsection{Notations and Problem Definition}
Throughout this paper vectors will be assumed to be column vectors and will be denoted by lower case letters, that is $\textbf{x}$. The transpose of $\textbf{x}$ will be denoted by $\textbf{x}^T$ and its Euclidean norm by $\norm{\textbf{x}} = \sqrt{\textbf{x}^T\textbf{x}}$. A multivariate Gaussian distribution of $\B{x}$ with mean $\bm{\mu}$ and covariance $\Sigma$ will be denoted using the notation $\B{x} \sim \mathcal{N}(\bm{\mu}, \Sigma)$. Matrices will be denoted by capital letters. The trace of a square matrix $M$ will be denoted by $tr(M)$. The identity matrix will be denoted by $I$ or $I_n$ when the dimension needs to be stressed. A diagonal matrix with diagonal elements $\lambda_1, \ldots, \lambda_n$ will be denoted by $diag(\lambda_1, \ldots, \lambda_n)$. Sets will be denoted using calligraphic capital letters like $\mathcal{S}$ or $\mathcal{R}$. Unless otherwise mentioned, subscripts on vectors/matrices will be used to denote time indexes and (whenever necessary) superscripts will be used to indicate the robot or the object that it refers to. For example, $\textbf{x}_k^i$ represents the state of robot $i$ at time instant $k$. The notation $P(\cdot)$ will be used to denote the probability of an event and the probability density function (pdf) will be denoted by $p(\cdot)$. While deriving the \textit{Belief Space Planning} (BSP) framework to incorporate object uncertainties we will mainly follow the notations and formalisms in~\cite{thrun2005book}.

We now formally define the problem that we tackle in this paper. Consider a robot operating in a partially-observable environment. The map of the environment is either known a priori or is built using a standard SLAM algorithm. At any time $k$, we denote the robot pose (or configuration) by $\B{x}_k\doteq(x_k, y_k, \theta_k)$, the acquired measurement from objects is denoted by $\B{z}_k$ and the applied control action is denoted as $\B{u}_k$. Note that by objects we refer to both the landmarks and the obstacles in the environment. We consider a standard motion model with Gaussian noise
\begin{equation}
\B{x}_{k+1} = f(\B{x}_k,\B{u}_k) + \B{w}_{k}\  ,  \ \B{w}_{k} \sim \mathcal{N}(0,R_{k})
\label{eq:odometry_model}
\end{equation}

\noindent
where $\B{w}_k$ is the random unobservable noise, modeled as a zero mean Gaussian. We note that modeling the random unobservable noise variables as Gaussians with zero mean is a common practice in robotics~\cite{thrun2005book}. The objects are detected through the robot's sensors and, assuming data association is known, the observation model can be written as  
\begin{equation}
\B{z}_k = h(\B{x}_k,O_k^i) + \B{v}_k \  ,  \ \B{v}_k \sim \mathcal{N}(0,Q_k)
\label{eq:measurement_model}
\end{equation}

\noindent
where $O_k^i$ is the detected $i$-th object and $\B{v}_k$ is the zero mean Gaussian noise. The function $h(\mathbf{x}_k,O_k^i)$ denotes the fact that at time $k$, the measurement $\mathbf{z}_k$ is obtained by observing the $i-th$ object $O_k^i$ from viewpoint (robot location) $\B{x}_k$. In the case of a laser-range finder the function $h$ could be defined as the distance between $\mathbf{x}_k$ and the location of the object (or any particular point on the object) $O_k^i$. If we consider the case of a camera, $h$  may be defined as a pinhole projection operator, projecting the object $O_k^i$ onto the image plane. Given the models in (\ref{eq:odometry_model}) and (\ref{eq:measurement_model}), in this paper we focus on two aspects. First, we consider the object uncertainties while localizing the robot. Second, we compute the exact probability of collision under obstacle uncertainty, which is modeled as a Gaussian distribution. Finally, we evaluate our approach in two simulation domains: a 2D mobile robot domain and a 2D manipulator domain. It is to be noted that for the manipulator domain we will be concerned with the collision avoidance of the manipulator's end-effector.  
\subsection{Related Work}
BSP has been researched extensively in the past with applications spanning a variety of areas including autonomous navigation, multi-modal planning, and active SLAM~\cite{van_den_berg2012IJRR, Kurniawati2016ISRR, agha_mohammadi2014IJRR, prentice2009IJRR, thomas2019ISRR, thomas2020STAIRS, kaelbling2013IJRR, pathak2018IJRR}.\cite{kaelbling2013IJRR} consider object uncertainty since they are planning in an unknown environment
and require several measurements to obtain confidence estimates of object locations. Thus they perform active perception, that is, to look for robot actions that enhances information to reduce the object uncertainty. This context is different from ours since we consider a known environment with object uncertainty and focus on active localization incorporating these uncertainties. In~\cite{pathak2018IJRR}, the concept of object uncertainty is commented upon (they call it scene uncertainty); however they do not show how it affects the state estimation. Dynamic environments are considered in~\cite{Kurniawati2016ISRR,agha_mohammadi2014IJRR} however the landmark/beacon locations are assumed to be known perfectly;~\cite{thomas2019ISRR, thomas2020STAIRS} also consider perfect landmark locations in the context of task and motion planning. Thus most active and passive localization-based approaches focus on robot state uncertainty and assume perfect knowledge about the location of the objects in the environment. However, in practice, the environment is seldom known with high certainty and hence providing formal guarantees for safe navigation is imperative.   

Patil \textit{et al.}~\cite{patil2012ICRA} estimate the probability of collision under robot state uncertainty by truncating the state distributions. In~\cite{bry2011ICRA}, future state distributions are predicted and the uncertainties are used to compute bounded collision probabilities. Lee \textit{et al.}~\cite{lee2013IROS} use sigma hulls\footnote{ Sigma hulls are convex hulls of the geometry of
individual robot links transformed according to the sigma
points in joint space~\cite{lee2013IROS}.} to formulate collision avoidance constraints in terms of the signed distance to the obstacles. Du Toit and Burdick~\cite{dutoit2011IEEE}, Park \textit{et al.}~\cite{park2018IEEE} compute the collision probability by marginalizing the joint distribution between the robot and obstacle location. The distributions are assumed to be Gaussian and the marginalization is computed with an indicator function that is true under the collision condition. However, since there is no closed-form solution to this formulation, an approximation is assumed. Furthermore, Park \textit{et al.} compute an upper bound for the collision probability. An approximation is computed using Monte Carlo Integration in~\cite{lambert2008ICCARV}, albeit computationally intensive. Another impressive work that uses Monte Carlo approach is \textit{Monte Carlo Motion
Planning} (MCMP)~\cite{janson2018ISRR}. This approach first solves a deterministic motion planning problem with inflated obstacles and then adjusts the inflation to compute a path that is exactly as safe as desired.

Linear chance constraints are used to compute bounded collision-free trajectories with dynamic obstacles in~\cite{zhu2019RAL}. Axelrod \textit{et al.}~\cite{axelrod2018IJRR} focus exclusively on obstacle uncertainty. They formalize a notion of ``shadows", which are the geometric equivalent of confidence intervals for uncertain obstacles. The shadows fundamentally give rise to loose bounds but the computational complexity of bounding the collision probability is greatly reduced. Uncertain obstacles are modelled as polytopes with Gaussian-distributed faces in~\cite{shimanuki2018WAFR}. Planning a collision-free path in the presence of ``risk zones" is considered in~\cite{salzman2017ICAPS} by penalizing the time spent in these risk zones. Risk contour maps which give the risk
information (uncertainties in location, size and geometry of obstacles) in uncertain environments are used in~\cite{jasour2019RSS} to obtain safe paths with bounded risks. A related approach for randomly moving obstacles is presented in~\cite{hakobyan2019RAL}. Formal verification methods have also been used to construct safe plans~\cite{ding2013ICRA, sadigh2016RSS}.

Most approaches discussed above compute the collision probability along a path by summing or multiplying the probabilities along different waypoints in the path. Boole's inequality is used to decouple the total probability in terms of individual waypoint probabilities. Such approaches tend to be overly conservative and rather than computing bounded collision probabilities along a path, the bound should be checked for each configuration along a path. Moreover, in most approaches, the collision probability computed along each waypoint is an approximation of the true value. On the one hand, such approximations can overly penalize paths and could gauge all plans to be infeasible. On the other hand some approximations can be lower\footnote{For example, the approach in~\cite{dutoit2011IEEE} computes a value lower than the actual when the robot state covariance is small.} than the true collision probability values and can lead to synthesizing unsafe plans.

\subsection{Contributions}
In this paper two main theoretical contributions are presented. First, we incorporate object uncertainties in the BSP planning framework and derive the resulting Bayes filter in terms of the prediction and measurement updates of the Extended Kalman Filter (EKF). The second is the computation of the probability of collision under environment uncertainty. We formulate the collision avoidance constraint as a quadratic form in random variables. This provides an exact expression for the collision probability in terms of a converging infinite series. A notion of safety is also formalized to compute configurations that satisfy the required collision probability bounds. 

We make the following assumptions: (1) the uncertainties are modelled using Gaussian distributions; (2) while formulating the collision constraint, we assume that the robot and obstacles have circular geometries. However, this is by no means a limitation and the approach can be extended to objects with different geometries by considering the configuration spaces.    

\section{Object Uncertainty}
In this Section, we focus on a BSP formulation that incorporates object uncertainties, that is, the viewpoints whence the objects can be observed are not precisely known. We define the \textit{object space} $\mathcal{O} = \{O^i| \text{$O^i$ is an object, and} \ 1 \leq i \leq |\mathcal{O}|\}$ to be the set of all objects in the environment. The motion (\ref{eq:odometry_model}) and observation (\ref{eq:measurement_model}) models can be written in a probabilistic framework as $p(\B{x}_{k+1}|\B{x}_k, \B{u}_k)$ and $p(\B{z}_k|\B{x}_k,O_k^i)$, respectively. Let us consider that at time $k$ the robot received a measurement $\B{z}_k$ which was originated by observing object $O^i_k$. Given an initial distribution $p(\B{x}_0)$, and the motion and observation models $p(\B{x}_{k+1}|\B{x}_k, \B{u}_k)$ and $p(\B{z}_k|\B{x}_k,O_k^i)$, the posterior probability distribution at time $k$ is the \textit{belief} $b[\B{x}_k]$ and can be written as $p(\B{x}_k|\B{z}_k, O_k^i, \B{z}_{0:k-1},\B{u}_{0:k-1}) $, where $O^i_k$ is the object observed at time $k$, $\B{z}_{0:k-1} \doteq \{\B{z}_0,...,\B{z}_{k-1}\}$ is the sequence of measurements up to $k-1$ and $\B{u}_{0:k-1} \doteq \{\B{u}_0,...,\B{u}_{k-1}\}$ is the sequence of controls up to $k-1$. Using Bayes rule and theorem of total probability, $b[\B{x}_k]$ can be expanded as 
\begin{multline}
p(\B{x}_k|\B{z}_k, O_k^i, \B{z}_{0:k-1},\B{u}_{0:k-1}) \\ = \eta_k p(\B{z}_k|\B{x}_k, O_k^i) p(O_k^i|\B{x}_k) \int_{\B{x}_{k-1}} p(\B{x}_k|\B{x}_{k-1},\B{u}_{k-1})b[\B{x}_{k-1}] 
\label{eq:posterior}
\end{multline}

\noindent where $\eta_k = 1/p(\B{z}_k|\B{z}_{0:k-1},\B{u}_{0:k-1})$ is the normalization constant and $b[\B{x}_{k-1}] \sim \mathcal{N} (\bm{\mu_{k-1}}, \Sigma_{k-1})$ is the belief at time $k-1$. The term $p(O_k^i|\B{x}_k)$ denotes the probability of observing the object $O_k^i$ from the pose $\B{x}_k$ and models the object uncertainty. Similarly, given an action $\B{u}_k$, the propagated belief can be written as
\begin{equation}
b[\bar{\B{x}_{k+1}}] = \int_{\B{x}_k} p(\B{x}_{k+1}|\B{x}_{k},\B{u}_{k})b[\B{x}_{k}] 
\end{equation}

Given the current belief $b[\B{x}_k]$ and the control $\B{u}_k$, the propagated belief parameters, that is, mean and covariance, can be computed using the standard EKF prediction as
\begin{equation}
\begin{split}
\bar{\bm{\mu}}_{k+1} & = f(\bm{\mu}_k, \bm{u}_k)\\
\bar{\Sigma}_{k+1}   & = F_{k} \Sigma_k F_{k}^T +R_{k}
\end{split}
\label{eq:predict}
\end{equation}

\noindent where $F_{k}$ is the Jacobian of $f(\cdot)$ with respect to $\B{x}_k$. To compute the posterior belief using EKF update equations, we first need to model the term $p(O_k^i|\B{x}_k)$. In this work we model the object distribution as a Gaussian distribution given by
\begin{equation}
p(O_k^i|x_k) \sim \mathcal{N}(\bm{\mu}_{O_k^i},\Sigma_{O_k^i}) 
\end{equation}

\noindent where $\bm{\mu}_{O_k^i}$ is the mean viewpoint/pose that corresponds to the maximum probability of observing $O_k^i$ and $\Sigma_{O_k^i}$ is the associated covariance. 

For convenience we state the probability density function (pdf) of multivariate Gaussian distributions. For $\B{x} \sim \mathcal{N}(\bm{\mu}, \Sigma)$ the pdf is of the form
\begin{equation}
p(\B{x}) = det\left(2\pi \Sigma \right)^{-\frac{1}{2}}\textrm{exp}\left(-\frac{1}{2}(\B{x} - \bm{\mu})^T \Sigma^{-1} (\B{x} - \bm{\mu}) \right)
\end{equation}

\noindent where $det(\cdot)$ denotes the determinant. Expanding the right hand side of (\ref{eq:posterior}), we have $b[\B{x}_{k+1}] = \eta'_k \int \text{exp}(-\mathcal{J}_{k+1})$, where $\eta'_k$ contains the non-exponential terms and $\mathcal{J}_{k+1}$ is given by
\begin{multline}
\mathcal{J}_{k+1} = \frac{1}{2}\left(\B{z}_{k+1} - h\left(\bar{\bm{\mu}}_{k+1}\right) - H_{k+1}\left(\B{x}_{k+1} - \bar{\bm{\mu}}_{k+1}\right)\right)^T
 \\ 
Q_{k+1}^{-1}\left(\B{z}_{k+1} - h\left(\bar{\bm{\mu}}_{k+1}\right) - H_{k+1}\left(\B{x}_{k+1} - \bar{\bm{\mu}}_{k+1}\right)\right) \\
 +  \frac{1}{2}(\bm{x}_{k+1} - \bm{\mu}_{O_{k+1}^i})^T\Sigma_{O_{k+1}^i}^{-1}(\B{x}_{k+1} -\bm{\mu}_{O_{k+1}^i})\\
 +  \frac{1}{2} (\B{x}_{k+1} - \bar{\bm{\mu}}_{k+1} )^T \bar{\Sigma}_{k+1}^{-1} (\B{x}_{k+1} - \bar{\bm{\mu}}_{k+1} )
 \label{eq:exponential_terms}
\end{multline}

\noindent where $H_{k+1}$ is the Jacobian of $h(\cdot)$ with respect to $\B{x}_{k+1}$. We note that when object uncertainty is not considered, the second term in~(\ref{eq:exponential_terms}) disappears and the results that we derive below reduce to that of the standard EKF update case. The parameters of this Gaussian can be obtained by taking the first and second derivatives of $\mathcal{J}_{k+1}$ with respect to $\B{x}_{k+1}$,
\begin{multline}
\frac{\partial \mathcal{J}_{k+1}}{\partial \B{x}_{k+1}} =  -H_{k+1}^TQ_{k+1}^{-1}\left(\B{z}_{k+1} - h(\bar{\bm{\mu}}_{k+1}) - \right. \\  \left.H_{k+1}(\B{x}_{k+1}  - \bar{\bm{\mu}}_{k+1})\right) + \Sigma_{O_{k+1}^i}^{-1}\left(\B{x}_{k+1} - \bm{\mu}_{O_{k+1}^i}\right) +  \\ \bar{\Sigma}_{k+1}^{-1} \left(\B{x}_{k+1} - \bar{\bm{\mu}}_{k+1} \right)
\label{eq:first_derivative}
\end{multline}
\begin{equation}
\frac{\partial^2 \mathcal{J}_{k+1}}{\partial \B{x}_{k+1}^2} = H_{k+1}^TQ_{k+1}^{-1}H_{k+1} + \Sigma_{O_{k+1}^i}^{-1} +\bar{\Sigma}_{k+1}^{-1}
\label{eq:second_derivative}
\end{equation}

\noindent The term (\ref{eq:second_derivative}) is the inverse of the covariance of $b[\B{x}_{k+1}]$~\cite{thrun2005book}, that is, 
\begin{equation}
\Sigma_{k+1} = \left( H_{k+1}^TQ_{k+1}^{-1}H_{k+1} + \Sigma_{O_{k+1}^i}^{-1} +\bar{\Sigma}_{k+1}^{-1}\right)^{-1}
\label{eq:cov}
\end{equation}

Since the mean of $b[\B{x}_{k+1}]$ is the value that minimizes $\mathcal{J}_{k+1}$, it is obtained by equating (\ref{eq:first_derivative}) to zero
\begin{multline}
H_{k+1}^TQ_{k+1}^{-1}\left(\B{z}_{k+1} - h\left(\bar{\bm{\mu}}_{k+1}\right) - H_{k+1}\left(\B{x}_{k+1} - \bar{\mu}_{k+1}\right)\right) \\ =  \Sigma_{k+1}^{-1}\left(\bm{\mu}_{k+1} - \bar{\bm{\mu}}_{k+1} \right)  - \Sigma_{O_{k+1}^i}^{-1}\left(\bm{\mu}_{O_{k+1}^i}- \bar{\bm{\mu}}_{k+1}\right) \\
\implies \bm{\mu}_{k+1} =  \bar{\bm{\mu}}_{k+1} + K_{k+1}\left(\B{z}_{k+1} - h\left(\bar{\bm{\mu}}_{k+1}\right)\right) \\
+ \Sigma_{k+1}\Sigma_{O_{k+1}^i}^{-1}\left(\bm{\mu}_{O_{k+1}^i}- \bar{\bm{\mu}}_{k+1}\right)
\label{eq:object_mu}
\end{multline}

\noindent where $K_{k+1}=\Sigma_{k+1} H_{k+1}^TQ_{k+1}^{-1}$ is the Kalman gain.  

As in the case of standard EKF, the gain $K_{k+1}$ can be transformed to an expression that does not depend on $\Sigma_{k+1}$, by post-multiplying with an identity matrix $I = AA^{-1}$, where
\begin{multline}
A = \\ \left(H_{k+1}\bar{\Sigma}_{k+1}\left(\bar{\Sigma}_{k+1} +\Sigma_{O_{k+1}^i} \right)^{-1}\Sigma_{O_{k+1}^i} H_{k+1}^T + Q_{k+1}\right)
\end{multline}

\noindent This gives
 \begin{multline}
K_{k+1} = 
\Sigma_{k+1} \left(H_{k+1}^TQ_{k+1}^{-1}H_{k+1}\bar{\Sigma}_{k+1}\left(\bar{\Sigma}_{k+1} + \Sigma_{O_{k+1}^i} \right)^{-1} \right. \\
\left. \Sigma_{O_{k+1}^i} H_{k+1}^T + H_{k+1}^T\right) A^{-1}
\label{eq:gain_derivation}
\end{multline}
\noindent In order to simplify the above expression for $K_{k+1}$, we first compute the inverse of the term 
\begin{equation}
\bar{\Sigma}_{k+1}\left(\bar{\Sigma}_{k+1} +\Sigma_{O_{k+1}^i} \right)^{-1}\Sigma_{O_{k+1}^i}
\end{equation}
\noindent The inverse is computed as
 \begin{multline}
\left(\bar{\Sigma}_{k+1}\left(\bar{\Sigma}_{k+1} +\Sigma_{O_{k+1}^i} \right)^{-1}\Sigma_{O_{k+1}^i} \right)^{-1} \\= \Sigma_{O_{k+1}^i}^{-1}\left(\bar{\Sigma}_{k+1} +\Sigma_{O_{k+1}^i} \right)\bar{\Sigma}_{k+1}^{-1} \\
 = \Sigma_{O_{k+1}^i}^{-1} \bar{\Sigma}_{k+1}\bar{\Sigma}_{k+1}^{-1} + \Sigma_{O_{k+1}^i}^{-1}\Sigma_{O_{k+1}^i}\bar{\Sigma}_{k+1}^{-1} \\
 = \Sigma_{O_{k+1}^i}^{-1} +\bar{\Sigma}_{k+1}^{-1}
\label{eq:inverse}
\end{multline}

\noindent Using (\ref{eq:inverse}) and (\ref{eq:cov}), the expression in (\ref{eq:gain_derivation}) simplifies to 
\begin{multline}
K_{k+1}  =\Sigma_{k+1} \left(H_{k+1}^TQ_{k+1}^{-1}H_{k+1}+ \Sigma_{O_{k+1}^i}^{-1} +\bar{\Sigma}_{k+1}^{-1}\right)\bar{\Sigma}_{k+1} \\
\left(\bar{\Sigma}_{k+1} +\Sigma_{O_{k+1}^i} \right)^{-1}\Sigma_{O_{k+1}^i} H_{k+1}^T \\
 \left(H_{k+1}\bar{\Sigma}_{k+1}\left(\bar{\Sigma}_{k+1} +\Sigma_{O_{k+1}^i} \right)^{-1}\Sigma_{O_{k+1}^i} H_{k+1}^T + Q_{k+1}\right)^{-1}\\
 = \bar{\Sigma}_{k+1}\left(\bar{\Sigma}_{k+1} +\Sigma_{O_{k+1}^i} \right)^{-1}\Sigma_{O_{k+1}^i} H_{k+1}^T \\  \left(H_{k+1}\bar{\Sigma}_{k+1}\left(\bar{\Sigma}_{k+1} +\Sigma_{O_{k+1}^i} \right)^{-1}\Sigma_{O_{k+1}^i} H_{k+1}^T + Q_{k+1}\right)^{-1}\\
 \vspace{-0.2cm}
\end{multline}

By treating the sum $\Sigma_{O_{k+1}^i}^{-1} +\bar{\Sigma}_{k+1}^{-1}$ in~(\ref{eq:cov}) as a single term and applying the matrix inversion lemma on the right hand side of~(\ref{eq:cov}) and further simplifying using the expression for the inverse computed in~(\ref{eq:inverse}), it can be shown that
\begin{equation}
\Sigma_{k+1} = \left(I - K_{k+1}H_{k+1}\right)\bar{\Sigma}_{k+1}\left(\bar{\Sigma}_{k+1} +\Sigma_{O_{k+1}^i} \right)^{-1}\Sigma_{O_{k+1}^i}
\label{eq:object_sigma}
\end{equation}

We note that when no object uncertainty is considered the update step of the standard EKF gives $\bm{\mu}_{k+1} =  \bar{\bm{\mu}}_{k+1} + K_{k+1}\left(\B{z}_{k+1} - h\left(\bar{\bm{\mu}}_{k+1}\right)\right)$ and $\Sigma_{k+1} = \left(I - K_{k+1}H_{k+1}\right)\bar{\Sigma}_{k+1}$. The additional term in~(\ref{eq:object_mu}) rightly adjusts the mean $\bm{\mu}_{k+1}$ accounting for the fact that the object location is uncertain. Similarly, the extra terms in~(\ref{eq:object_sigma}) account for the object uncertainty and scale the posterior covariance accordingly.  

\section{Collision Probability}
Let $\mathcal{R}$ represent the set of all points occupied by a rigid-body robot at any given time. Thus, $\mathcal{R}$ represents the collection of points that form the rigid-body robot. Similarly, let $\mathcal{S}$ represent the set of all points occupied by a rigid-body obstacle. A collision occurs if $\mathcal{R} \cap \mathcal{S} \neq \{\phi\}$ and we denote the probability of collision as $P\left(\mathcal{R} \cap \mathcal{S} \neq \{\phi\}\right)$. In this work we assume circular geometries for $\mathcal{R}$ and $\mathcal{S}$ with radii $r_1$ and $s_1$, receptively and we denote the center of mass of the robot and the obstacle by $\B{x}_k$ and $\B{s}$, receptively. By abuse of notation we will use $\B{x}_k$ and $\B{s}$ equivalently to $\mathcal{R}$ and $\mathcal{S}$. The collision condition will be written in terms of the center of mass as $\mathcal{C}_{\B{x}_k,\B{s}}: \mathcal{R} \cap \mathcal{S} \neq \{\phi\}$. It is noteworthy that both $\B{x}_k$ and $\B{s}$ are not known precisely but can only be estimated probabilistically, as seen in the previous section. At this point we would like to stress the fact that the concepts and the derivations herein are valid for any 2D rigid-body robot. A mobile robot may be represented by a minimum area enclosing circle. In the case of a 2D manipulator robot each link can be approximated by bounding circles that tightly enclose the link. For such robots, the collision with an obstacle has to be checked for each bounding circle. For example, consider a manipulator robot with $l$ bounding circles. Then the collision condition for the $i-$th circle ($1\leq i \leq l$) is given by $\mathcal{C}_{\B{x}_k^i,\B{s}}$, where $\B{x}_k^i$ is the center of the $i-$th circle. 

Let us now consider an obstacle at any given time instant, distributed according to the Gaussian $\B{s} \sim \mathcal{N}\left(\bar{\B{s}},\Sigma_s\right)$, where $\bar{s}$ represents the mean and $\Sigma_s$ the uncertainty in the estimation of the object. Given the belief at time $k$, that is, $b[\B{x}_k]$, the probability of collision is given by
\begin{equation}
P\left(\mathcal{C}_{\B{x}_k,\B{s}}\right) = \int_{\B{x}_k} \int_{\B{s}} I_c(\B{x}_k,\B{s})p(\B{x}_k,\B{s})
\label{eq:collision_prob}
\end{equation}
where $\mathcal{C}_{\B{x}_k,\B{s}}$ as defined above represents the fact that robot configuration $\B{x}_k$ and its collision with obstacle at location $\B{s}$ is considered, and $I_c$ is an indicator function defined as
\begin{equation}
   I_c(\B{x}_k,\B{s})= 
   \begin{cases}
     1 \ &\text{if} \ \mathcal{R} \cap \mathcal{S} \neq \{\phi\} \\
     0 \ &\text{otherwise}.
   \end{cases}
\end{equation}

Du Toit and Burdick~\cite{dutoit2011IEEE}, Park \textit{et al.}~\cite{park2018IEEE} approximate the integral in (\ref{eq:collision_prob}) as $Vp(\B{x}_k,\B{s})$, where $V$ is the 2D footprint (area) occupied by the robot. For this approximation, in~\cite{dutoit2011IEEE} it is assumed that the robot radius $\varepsilon$ is negligible and a point obstacle is considered for this derivation. 

To do away with this approximation, we formulate the above problem by considering an alternative approach. Since the robot and obstacle are assumed to be spherical objects, the collision constraint can be written as
\begin{equation}
\norm{\B{x}_k - \B{s}}^2 \leq (r_1+s_1)^2
\label{eq:coll_condition}
\end{equation} 

\noindent
where $\B{x}_k$ and $\B{s}$ are the random vectors that denote the robot and obstacle pose respectively. Here, $\B{x}_k$ and $\B{s}$ corresponds to the body-fixed frames in the global frame. As noted before, the two random vectors in~(\ref{eq:collision_prob}) are distributed according to $\B{s} \sim \mathcal{N}\left(\mathbf{\bar{\B{s}}},\Sigma_s\right)$ and $\B{x}_k \sim \mathcal{N}\left(\bm{\mu}_k,\Sigma_{k} \right)$. Let us denote by $\B{w} = \B{x}_k -\B{s}$, the difference between the two random variables. Then we know that $\B{w} $ is also a Gaussian, distributed as $\B{w}  \sim \mathcal{N} \left(\bm{\mu}_k - \bm{\bar{s}}, \Sigma_k + \Sigma_s \right)$. The collision constraint can now be written as
\begin{equation}
\B{v} = \norm{\B{w}}^2 = \B{w}^T\B{w} \leq (r_1+s_1)^2
\label{eq:collision}
\end{equation}
where $\B{v}$ is a random vector distributed according to the squared $L_2$-norm of $\B{w}$. Now, given the probability density function (pdf) of $\B{v}$, the collision constraint in~(\ref{eq:coll_condition})reduces to solving the integral
\begin{equation}
P\left(\mathcal{C}_{\B{x}_k,\B{s}}\right) = \int_{0}^{(r_1+s_1)^2} p(v)
\end{equation}
where $p(v) = P_{\B{v}}(\B{v} = v)$ is the pdf of $\B{v}$. It is noteworthy that the above expression is the cumulative distribution function (cdf) of $\B{v}$, which is defined as $F_{\B{v}}\left((r_1+s_1)^2\right) = P\left(\B{v} \leq (r_1+s_1)^2\right)$.

\subsection{Quadratic Form in Random Variables}
A quadratic form in random variables is defined as~\cite{provost1992book},
\begin{definition}
Let $\B{x} = \left(x_1, \ldots,x_n \right)^T$ denote a random vector with mean $\bm{\mu} = \left(\mu_1,\ldots,\mu_n\right)^T$ and covariance matrix $\Sigma$. Then the quadratic form in the random variables $x_1, \ldots,x_n $ associated with an $n \times n$ symmetric matrix $A = (a_{ij})$ is

\begin{equation}
Q(\B{x}) = Q(x_1, \ldots,x_n) = \B{x}^TA\B{x} = \sum_{i=1}^{n}\sum_{j=1}^n a_{ij}X_iX_j
\end{equation}
\end{definition}

Let us define $\B{y} = \Sigma^{-\frac{1}{2}}\B{x}$ and define a random vector $\B{z} = \left(\B{y} - \Sigma^{-\frac{1}{2}}\bm{\mu}\right)$. The resulting distribution of $\B{z}$ is thus zero mean with covariance being the identity matrix. Thus the quadratic form becomes
\begin{equation}
Q(\B{x})  = \left(\B{z} + \Sigma^{-\frac{1}{2}}\bm{\mu}\right)^T\Sigma^{\frac{1}{2}}A\Sigma^{\frac{1}{2}}\left(\B{z} + \Sigma^{-\frac{1}{2}}\bm{\mu}\right)
\end{equation}

Suppose there exists an orthogonal matrix $P$, that is, $PP^T = I$ which diagonalizes $\Sigma^{\frac{1}{2}}A\Sigma^{\frac{1}{2}}$, then $P^T\Sigma^{\frac{1}{2}}A\Sigma^{\frac{1}{2}}P = \textrm{diag}\left(\lambda_1,\ldots,\lambda_n\right)$, where $\lambda_1,\ldots,\lambda_n$ are the eigenvalues of $\Sigma^{\frac{1}{2}}A\Sigma^{\frac{1}{2}}$. The quadratic form can now be written as
\begin{equation}
\begin{split}
Q(\B{x}) & = \left(\B{z} + \Sigma^{-\frac{1}{2}}\bm{\mu}\right)^T\Sigma^{\frac{1}{2}}A\Sigma^{\frac{1}{2}}\left(\B{Z} + \Sigma^{-\frac{1}{2}}\bm{\mu}\right)\\
& = \left(\B{u} + \B{b}\right)^T\textrm{diag}\left(\lambda_1,\ldots,\lambda_n\right)\left(\B{u} + \B{b}\right)
\end{split}
\label{eq:quad_form}
\end{equation}

\noindent
where $\B{u} = P^T \B{z} = (u_1,\ldots,u_n)^T$ and $\B{b} = P^T \Sigma^{-\frac{1}{2}}\bm{\mu} = (b_1,\ldots,b_n)^T$. The expression in (\ref{eq:quad_form}) can be written concisely,
\begin{equation}
Q(\B{x}) = \B{x}^TA\B{x} = \sum_{i=1}^n \lambda_i (u_i + b_i)^2
\end{equation}

\begin{theorem}
The cdf of $Q(\textnormal{\B{x}}) = \textnormal{\B{y}} = \textnormal{\B{x}}^TA\textnormal{\B{x}}$ with $A = A^T > 0, \textnormal{\B{x}} \sim \mathcal{N}(\bm{\mu},\Sigma), 
\Sigma > 0$ is 

\begin{equation}
F_{\textnormal{\B{y}}}(y) = P(\textnormal{\B{y}}\leq y) = \sum_{k=0}^{\infty}(-1)^k c_k \frac{y^{\frac{n}{2} + k}}{ \Gamma\left(\frac{n}{2} +k +1\right)}
\label{eq:collision_probability}
\end{equation}

and its pdf is given by 

\begin{equation}
p_{\textnormal{\B{y}}}(y) = P(\textnormal{\B{y}} = y) = \sum_{k=0}^{\infty}(-1)^k c_k \frac{y^{\frac{n}{2} + k -1}}{ \Gamma\left(\frac{n}{2} +k \right)}
\label{eq:pdf}
\end{equation}
\label{theorem1}
\end{theorem}

\noindent
where $\Gamma$ denotes the gamma function and
\begin{equation*}
\begin{split}
& c_0 = \textrm{exp}(-\frac{1}{2}\sum\limits_{i=1}^{n}b_i^2)\prod_{i=1}^n\left(2\lambda_i\right)^{-\frac{1}{2}} \\
& c_k = \frac{1}{k}\sum\limits_{i=0}^{k-1}d_{k-i}c_i \\
& d_k = \frac{1}{2}\sum\limits_{i=1}^n \left(1-kb_i^2\right)\left(2\lambda_i\right)^{-k}
\end{split}
\end{equation*}

The proof of the above theorem is beyond the scope of this paper and we refer the interested readers to~\cite{provost1992book}. It is easily seen that the left hand side of (\ref{eq:collision}), is in the quadratic form $Q(\B{y})$ with $A = I$, the identity matrix. Thus the collision probability can be computed from (\ref{eq:collision_probability}) as 
\begin{equation}
  P\left(\mathcal{C}_{\B{x}_k,\B{s}}\right) =  F_{\B{y}}\left((r_1+s_1)^2\right)
\end{equation}

\subsection{Convergence and Truncation Error}
In this section we will prove the convergence the infinite series in~(\ref{eq:collision_probability}) and~(\ref{eq:pdf}). Note that the series expansion of the pdf in Theorem~\ref{theorem1} is of the form
\begin{equation}
p_{\B{y}}(y) = \sum\limits_{k=0}^{\infty} c_kh_k(y)
\end{equation}  
\noindent From~\cite{kotz1967AMS} we have the following lemma.

\begin{lemma}
Let $\{h_k\}_0^{\infty}$ be a sequence of measurable complex valued functions on $[0, \infty]$ and $\{c_k\}_0^{\infty}$ be a sequence of complex numbers such that

\begin{equation}
\sum_{k=0}^{\infty} |c_k||h_k(y)| \leq \alpha e^{(\beta y)} \ \textrm{for} \ y \in [0, \infty]  
\label{eq:convergence}
\end{equation}

where $\alpha$, $\beta$ are real constants. Then $L\left(h_k(y)\right)$ and $L(p_{\B{y}}(y))$ exist for $Re(s) > \beta$, and

\begin{equation}
L(p_{\B{y}}(y)) = \sum_{k=0}^{\infty} c_k L(h_k(s))
\end{equation}
\label{lemma1}
\end{lemma}

\noindent where $L(\cdot)$ denotes the Laplace transform. Let us now define the term $M(\theta)$ such that
\begin{equation}
M(\theta) = \sum_{k=0}^{\infty} c_k \theta^k
\end{equation}
\noindent where the infinite series is a uniformly convergent series for $\theta$ in some region with $M(\theta) > 0$. Let the Laplace transform of $h_k(y)$ be the form $L(h_k(y)) = \xi(s)\eta^k(s)$, where, for $Re(s) > \beta$ with $\beta$ being a real constant, $\xi(s)$ is a non-vanishing analytic function and $\eta(s)$ is an analytic function with an inverse function $\eta(\zeta(\theta)) = \theta$. For $h_k(y)$ in (\ref{eq:pdf}), we have, $\xi(s) = (2s)^{-n/2}$, $\eta(s) = -(2s)^{-1}$ and $\zeta(\theta) = -(2 \theta)^{-1}$. Now let us define,
\begin{equation}
\begin{split}
M(\theta) & = \left(L(p_{\B{y}}) \circ \zeta / \xi \circ \zeta \right)(\theta) = \sum_{k=0}^{\infty} c_k \theta^k
\end{split}
\end{equation}
\noindent where $\circ$ denotes function composition. Using Cauchy's inequality, we get
\begin{equation}
|c_k| \leq \frac{m(\rho)}{\rho^k}, \ \ m(\rho) = \textrm{max}_{|\theta| = \rho} |M(\theta)|
\label{eq:cauchy}
\end{equation}

\noindent Since $h_k(y)$ is bounded and, using (\ref{eq:cauchy}), the condition (\ref{eq:convergence}) in Lemma~\ref{lemma1} is satisfied and the series $p_{\B{y}}(y)$ converges uniformly in every bounded interval of $y >0$. As a result, integrating $p_{\B{y}}(y)$ term-by-term, the obtained series $F_{\B{y}}(y)$ is uniformly convergent in every bounded interval of $y >0$. 

If the series in (\ref{eq:pdf}) is truncated after $N$ terms, the truncation error is
\begin{equation}
e(N) = \sum_{k=N+1}^{\infty}|c_kh_k(y)| = \left|\sum_{k=N+1}^{\infty} c_k \frac{y^{\frac{n}{2} + k -1}}{ \Gamma\left(\frac{n}{2} +k \right)}\right|
\end{equation}
Using (\ref{eq:cauchy}), an upper bound for the truncation error can hence be obtained as
\begin{equation}
e(N) \leq \frac{m(\rho)}{\rho^k}\left|\sum_{k=N+1}^{\infty} \frac{y^{\frac{n}{2} + k -1}}{ \Gamma\left(\frac{n}{2} +k \right)}\right|
\end{equation}

\noindent
where the summation term can be further simplified using the gamma function identity, $\forall \varsigma > 0, \ \Gamma(\varsigma + 1) = \varsigma\Gamma(\varsigma)$, giving
\begin{equation}
e(N) \leq m(\rho)\left(\Gamma\left(\frac{n}{2} \right)N!\right)^{-1}(\frac{y}{2})^{\frac{n}{2} -1}(\frac{y}{2\rho})^{N+1}\text{exp}(\frac{y}{2\rho})
\end{equation}
\\

\noindent
The truncation error for (\ref{eq:collision_probability}) is obtained in a similar manner,
\begin{equation}
\displaystyle
E(N) \leq m(\rho)\left(\Gamma \left(\frac{n}{2}\right)(N+1)!\right)^{-1}(\frac{y}{2})^{\frac{n}{2}}(\frac{y}{2\rho})^{N+1}\text{exp}(\frac{y}{2\rho})
\label{eq:truncation}
\end{equation}  

The expression for $m(\rho)$ is obtained from~\cite{kotz1967AMS2},
\begin{equation}
m(\rho) = \prod_{j=1}^n \lambda_j^{-\frac{1}{2}} \text{exp} \left(-\frac{1}{2} \sum_{j=1}^n \frac{b_j^2\lambda_j}{ \lambda_j + \rho}\right)\prod_{j=1}^n (1 - \frac{\rho}{\lambda_j})^{-\frac{1}{2}}
\label{eq:positive}
\end{equation}

The expression in~(\ref{eq:positive}) is valid only if $\rho < \lambda_j$~\cite{kotz1967AMS} and hence $\rho < \textrm{min} \ \lambda_j$. Thus we have $m(\rho) \rightarrow 0$ with $\sum_{j=1}^n b_j^2 \rightarrow \infty$. The larger the distance from the obstacles and the higher the certainty in the robot and obstacle positions, the greater is the $b_j$ (see~\ref{eq:quad_form}) value. In such scenarios, convergence is often attained within the first few terms of the series. For a given robot configuration and obstacle parameters, we see that the only varying term in (\ref{eq:truncation}) is $(y/2\rho)^{N+1}/(N+1)!$ which depends on $\lambda_j$'s, that is the eigenvalues of  $\Sigma_k + \Sigma_s $. Clearly, at time instant $k$, the parameter that influences the convergence is the degree of uncertainty in both the robot and obstacle location, that is, $\Sigma_k + \Sigma_s $. 

\begin{figure}[t]
\centering
  \subfloat[Configuration A]{\includegraphics[scale=0.27]{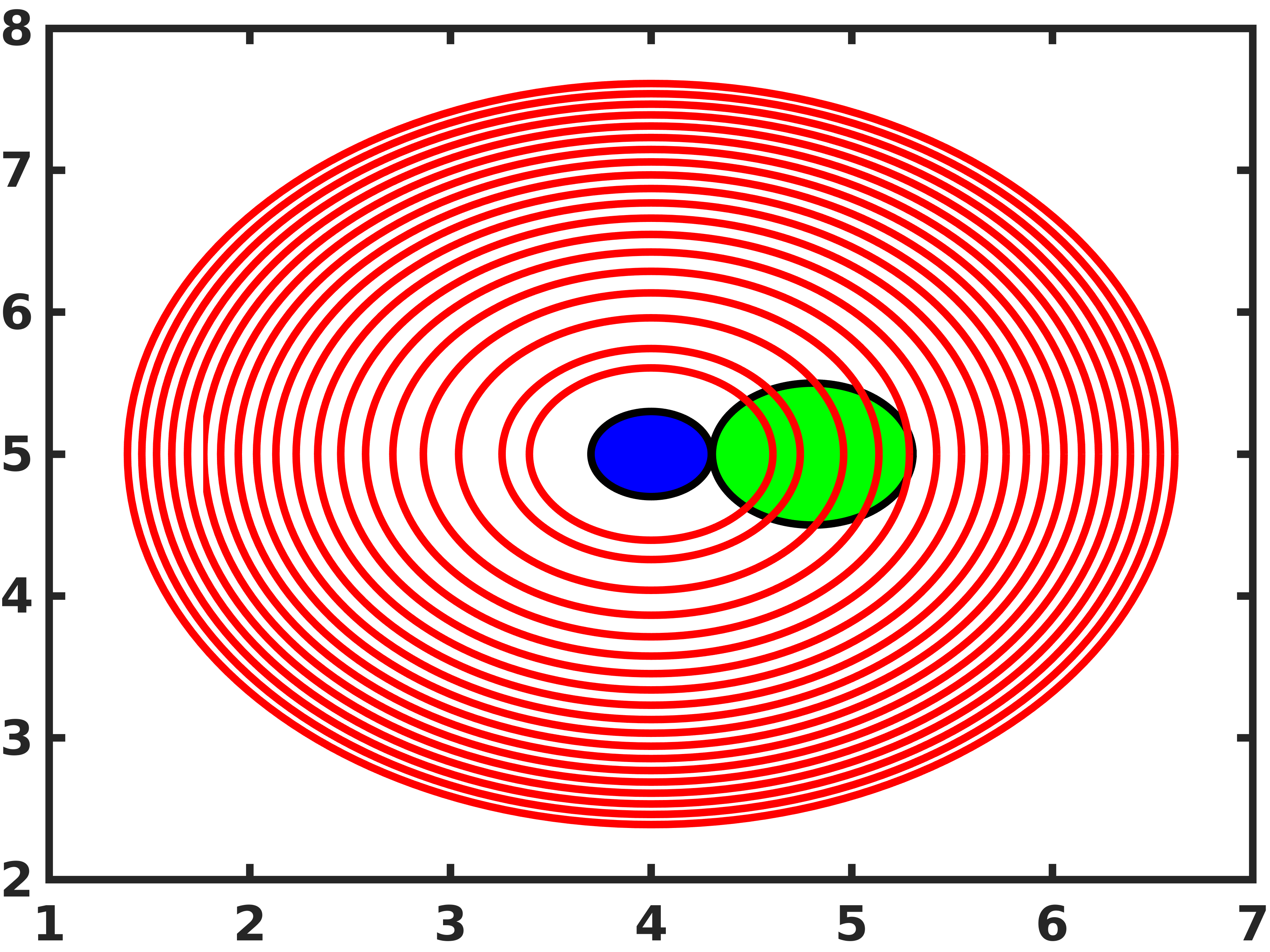}} 
  \hspace{0.2cm}
  \subfloat[Collision probability evolution]{\includegraphics[scale=0.27]{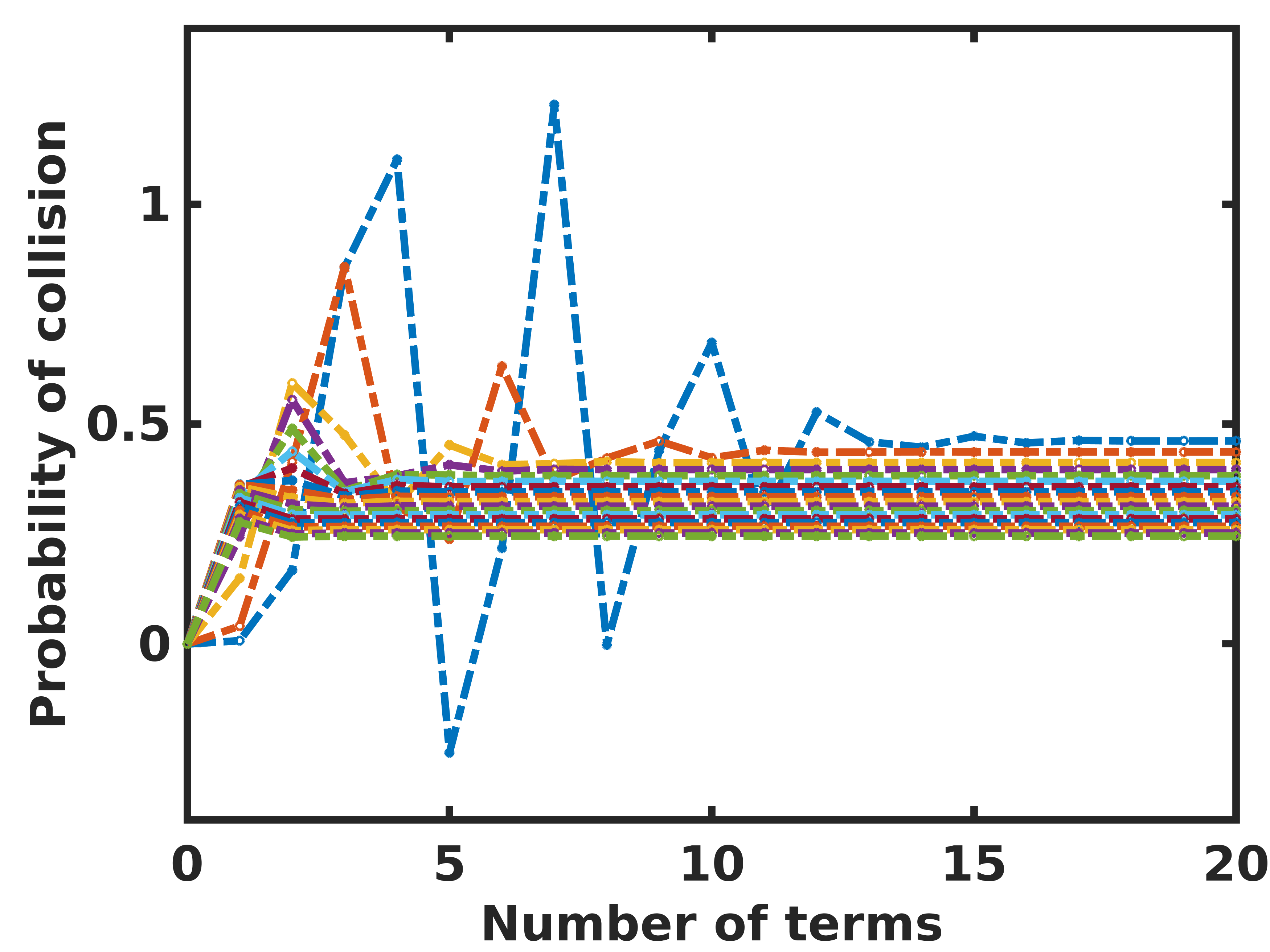}}\\
    \subfloat[Configuration B]{\includegraphics[scale=0.27]{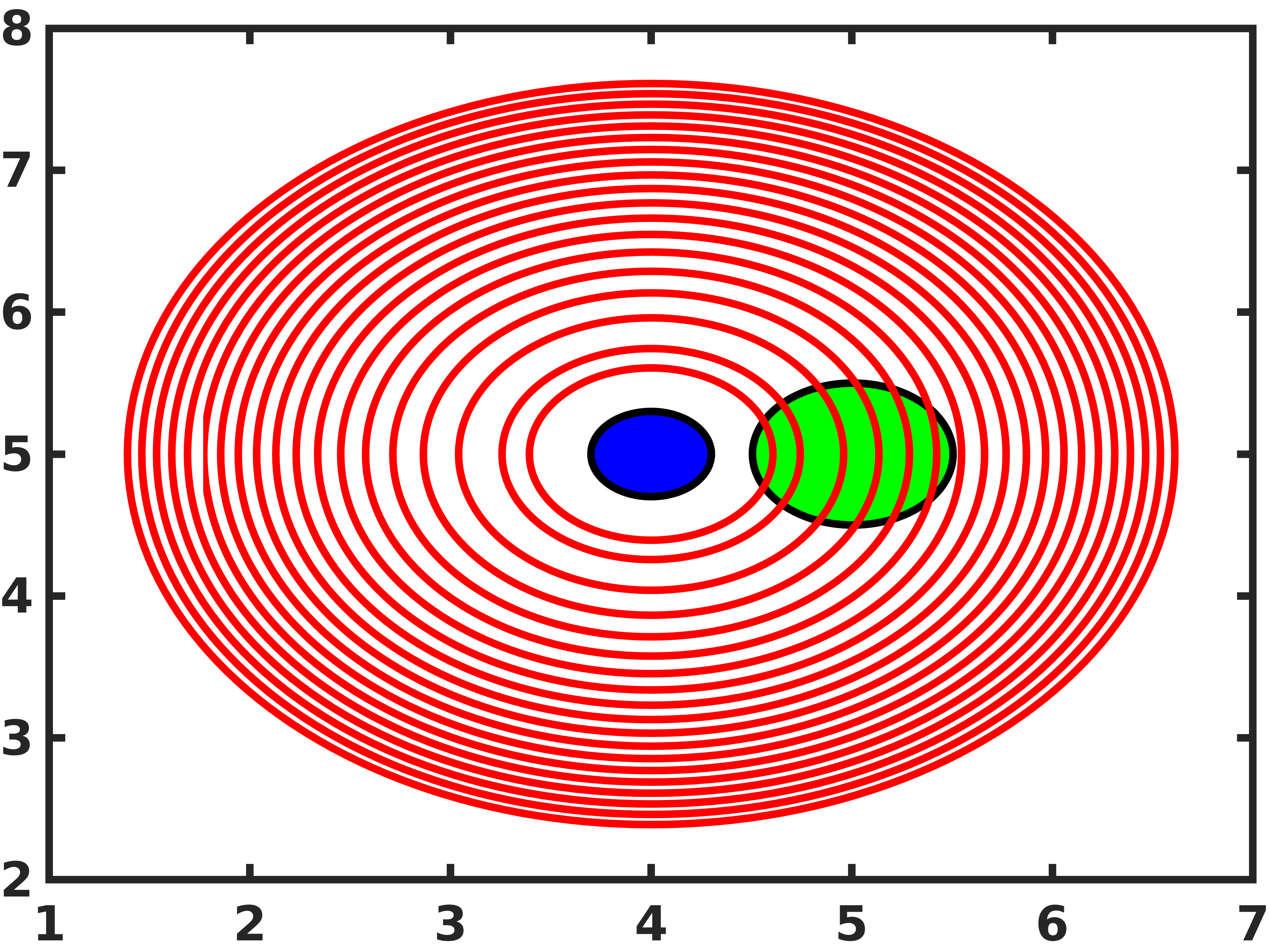}}
    \hspace{0.2cm} 
  \subfloat[Collision probability evolution]{\includegraphics[scale=0.27]{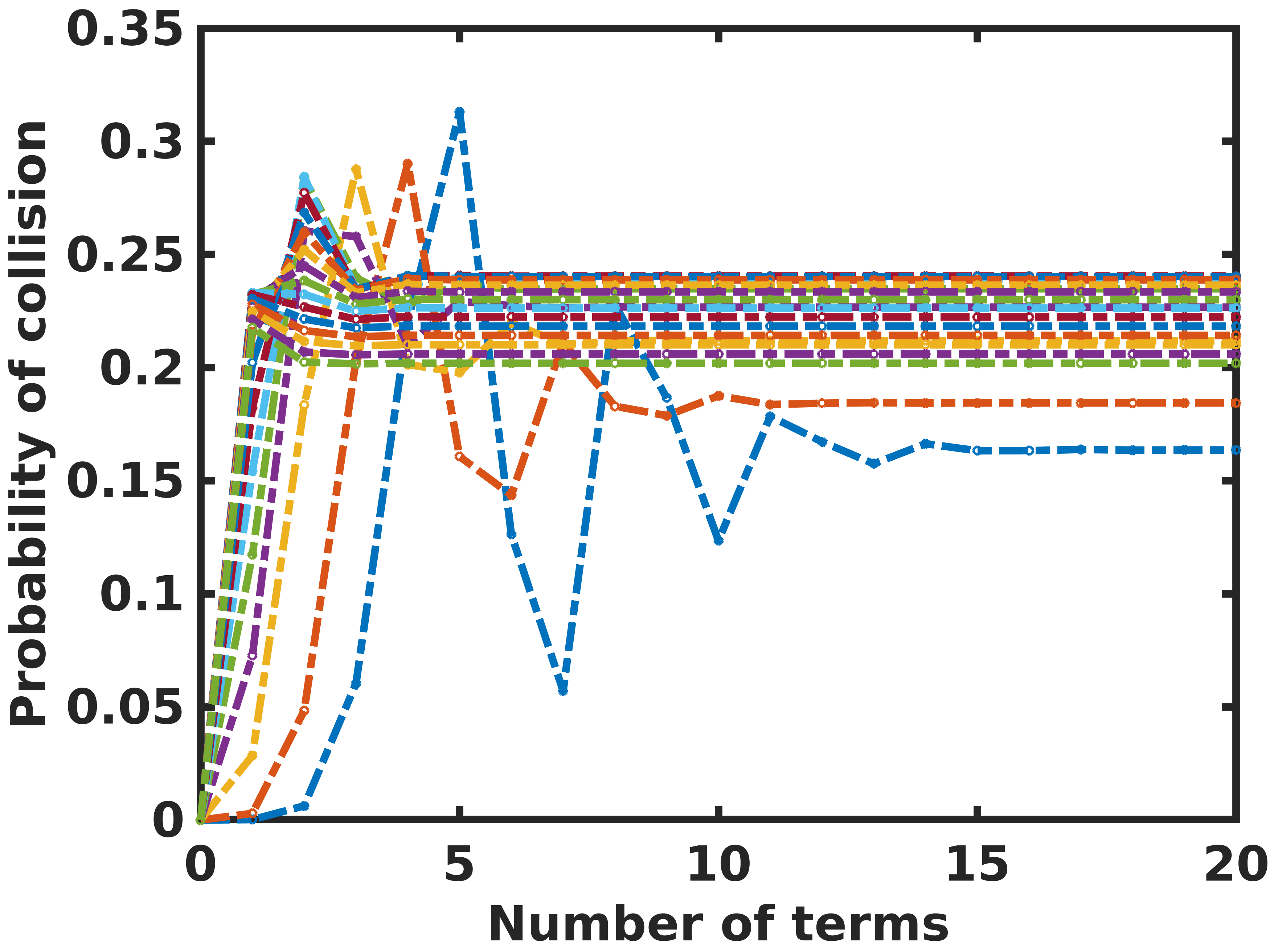}}\\
    \subfloat[Configuration C]{\includegraphics[scale=0.27]{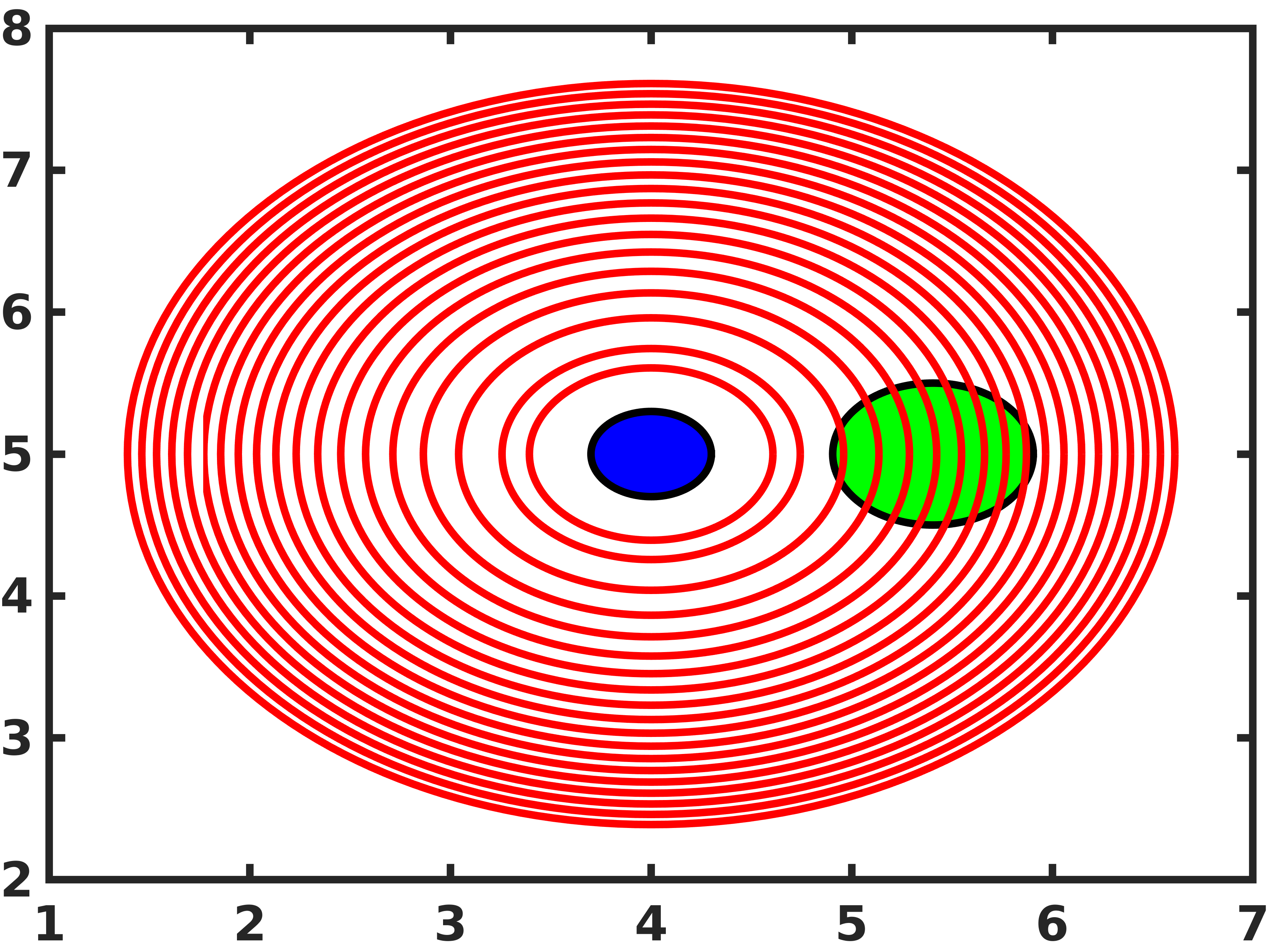}} 
    \hspace{0.2cm}
  \subfloat[Collision probability evolution]{\includegraphics[scale=0.27]{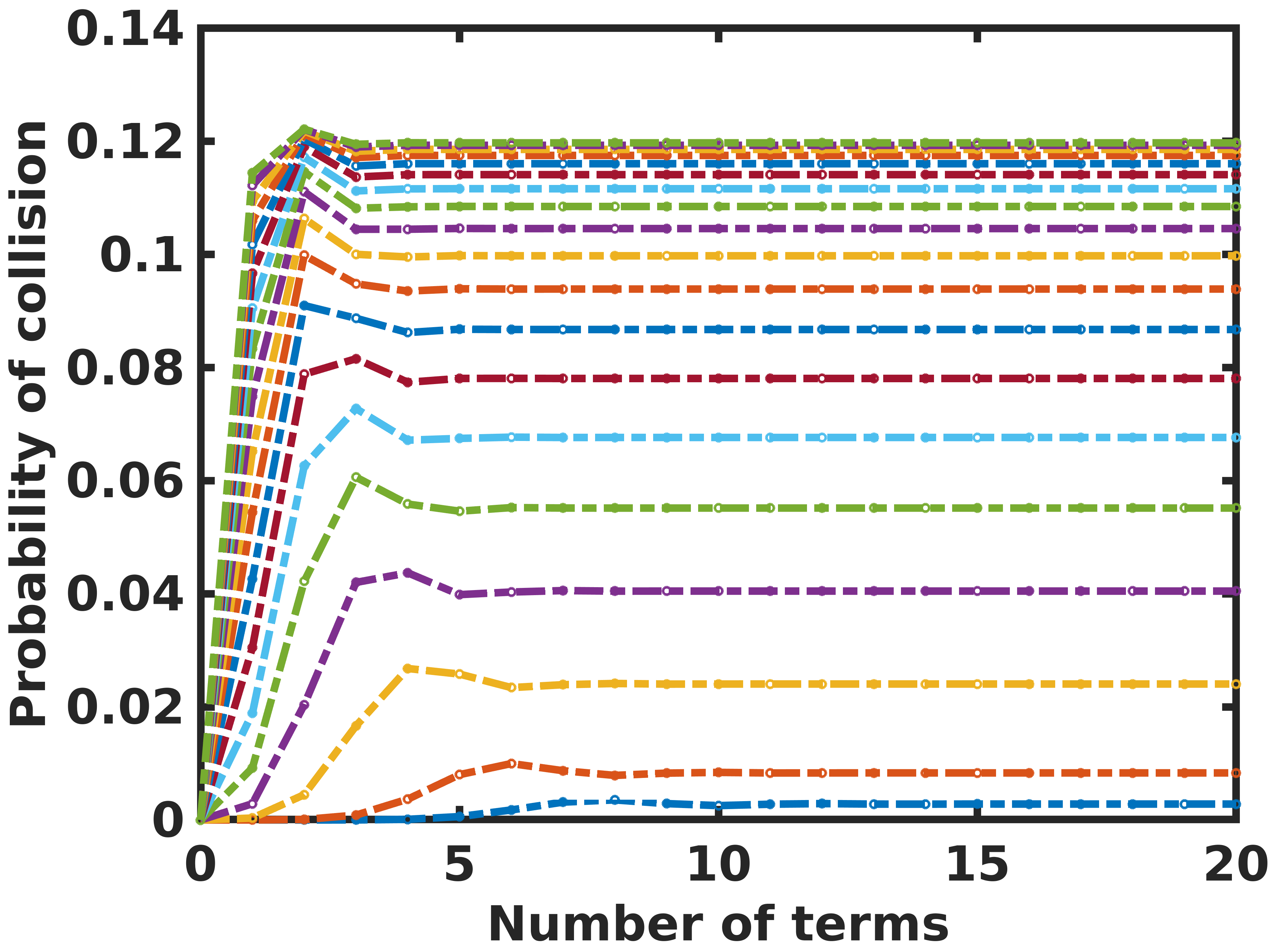}}\\
   \subfloat[Configuration D]{\includegraphics[scale=0.27]{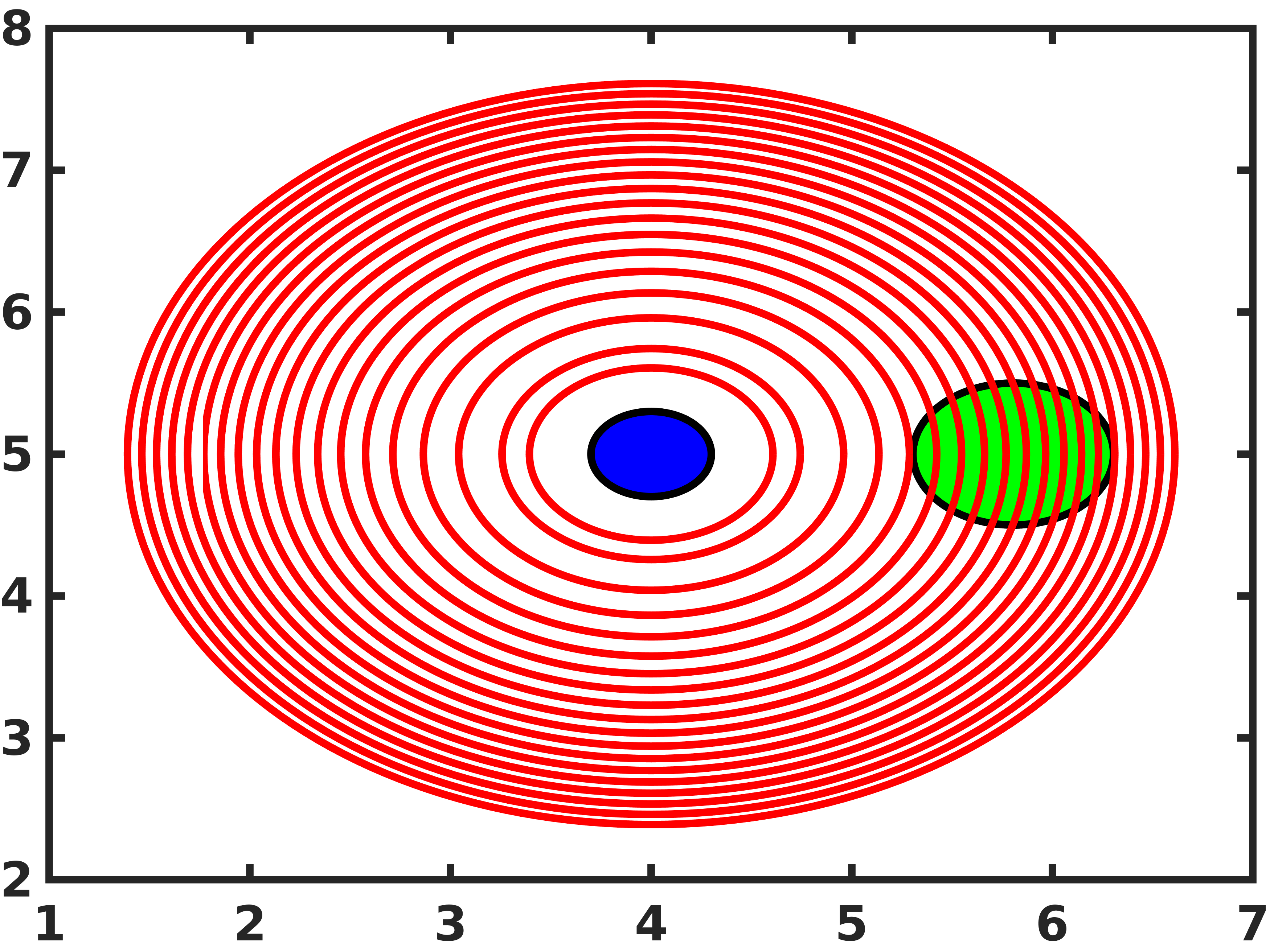}} 
   \hspace{0.2cm}
  \subfloat[Collision probability evolution]{\includegraphics[scale=0.27]{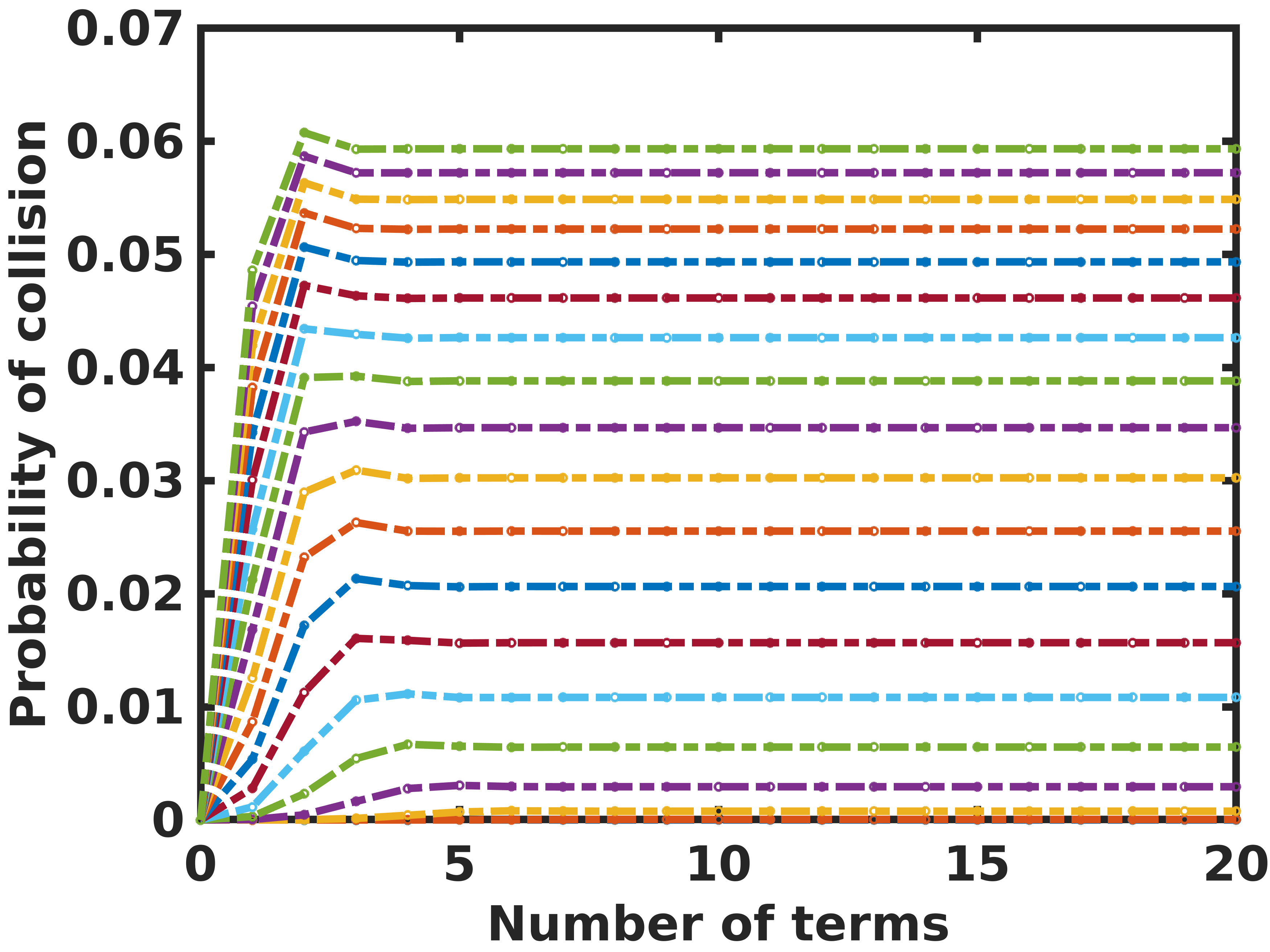}}
  \caption{Different configurations for a robot of radius 0.3$m$ and obstacle of radius 0.5$m$. For each configuration the evolution of probability of collision is plotted for different covariances. In each of the 4 configurations, maximum terms for convergence is for the minimum covariance of $diag(0.04,0.04)$.}
  \label{fig:convergence}
\end{figure}

The convergence is visualized for different configurations in Fig.~\ref{fig:convergence}. The blue and green circles represent a robot and an obstacle, respectively. The red ellipses corresponds to the 3$\sigma$ uncertainties for different covariances $diag(0.04,0.04),\ diag(0.08,0.08),\ \ldots,\ diag(0.74,0.74)$. In Fig.~\ref{fig:convergence}(a) the robot and the obstacle are touching each other. For each of these covariances, the number of terms for convergence is shown in Fig.~\ref{fig:convergence}(b). The worst case corresponds to the covariance of $diag(0.04,0.04)$, requiring 16 terms for convergence (dashed blue line with spikes in Fig.~\ref{fig:convergence}(b)). In Fig.~\ref{fig:convergence}(c) the distance between the robot and the obstacle is increased by 0.2$m$ and the covariance $diag(0.04,0.04)$ needed 12 terms for convergence. The distances are further increased by 0.4$m$ and 0.8$m$ in Fig.~\ref{fig:convergence}(e), (g) and their worst case convergences are 9 and 5 respectively as seen in Fig.\ref{fig:convergence}(f), (h). The number of terms for worst case convergence that corresponds to covariance $diag(0.04,0.04)$ and the respective time for collision probability computation are shown in Table~\ref{table1}.  
\begin{table}
\centering
\scalebox{0.85}{
\begin{tabular}{ |c|c|c| } 
 \hline
 Configuration & Terms for convergence & Computation time (s) \\
 \hline 
 A & 16 & 0.0412 $\pm$ 0.0086 \\ 
  \hline 
 B & 12 & 0.0044 $\pm$ 0.0041 \\ 
 \hline
 C & 9  & 0.0008 $\pm$ 0.0003 \\
 \hline
 D & 5  & 0.0004 $\pm$ 0.0002 \\
 \hline
\end{tabular}}
 \caption{The maximum number of terms required for convergence and the corresponding collision probability computation time. The values correspond to the covariance $diag(0.04,0.04)$ for each of the configurations.}
 \label{table1}
\end{table}

\subsection{Safe Configuration}
 
In the presence of perception and motion uncertainty, providing safety guarantees for robot motion is imperative. Let us assume that the obstacle position is known with high certainty as a result of perfect sensing. However, since the true state of the robot is not known and only a distribution of these states can be estimated, collision checking has to be performed for this distribution of states. Moreover, in practice, the observations are noisy and this renders the estimated obstacle location (and shape) uncertain. Hence, this uncertainty should be taken into account while considering collision avoidance. 

Given a robot configuration $\B{x}_k$, we define the following notion of $\epsilon-$safe configuration.

\begin{definition}
A robot configuration $\B{x}_k$ is an $\epsilon-$safe configuration with respect to an obstacle location $\B{s}$, if the probability of collision is such that $P\left(\mathcal{C}_{\B{x}_k,\B{s}}\right) \leq 1 - \epsilon$.
\end{definition}
For example, a $0.99-$safe configuration implies that the probability of this configuration colliding with the obstacle is at most $0.01$. We use the sampling based Probabilistic Roadmap (PRM)~\cite{kavraki1996IEEE} to compute motion plans. As a result we can only guarantee probabilistic completeness for returning $\epsilon-$safe configurations since the PRM motion planner is probabilistically complete~\cite{karaman2011IJRR}, that is the probability of failure decays to zero exponentially with the number of samples used in the construction of the roadmap. The failure to find an $\epsilon-$safe configuration might be because such a configuration indeed does not exist or simply because there were not enough samples.

%
%
%
\subsection{Complexity Analysis} 

It is known that for $m$ nodes, the computational complexity of PRM is $O(m\log m)$~\cite{karaman2011IJRR}. First let us consider the case of belief space planning over the PRM graph, without computing the collision probabilities. Finding a trajectory to the goal requires performing Bayesian (EKF) update operations. This basically involves performing matrix operations--- matrix multiplication and inversion of matrices. For a state of dimension $n$, the covariance matrix is of dimension $O(n^2)$. Therefore, each step of the Bayesian update has a complexity of $O(n^3)$. If $T$ denotes the number of time steps in the trajectory, then the overall computational complexity is $O(n^3T)$. Let us now analyze the complexity of collision probability computation. From (\ref{eq:truncation}) we see that for each iteration, the truncation error varies with $(y/2\rho)$. Therefore, to achieve $E(N) \leq \delta$, for an $\epsilon-$safe configuration, $k = O\left(\log \frac{\delta \rho}{y(1-\epsilon)}\right)$ iterations are required. We note that for each obstacle, the runtime is increased by this factor.

\section{Cost Function} 
At each time instant the robot is required to minimize its control usage and proceed towards the goal $\B{x}^g$, while minimizing its state uncertainty. We quantify the state uncertainty by computing the trace of the marginal covariance of the robot position. As a result, we have the following cost function
\begin{equation}
c \doteq \norm{\xi(\B{u}_k)}^2_{M_u} + \norm{\B{x}_k - \B{x}^g}^2_{M_g} + tr\left(\norm{M_{\Sigma}}^2_{\Sigma_k}\right) + M_CP(\mathcal{C})
\label{eq:cost}
\end{equation}  

\noindent where $\norm{x}_S = \sqrt{x^TSx}$ is the Mahalanobis norm, $M_u, M_g, M_C$ are weight matrices and $\xi(\B{u}_k)$ is a function that quantifies control usage. The choice of weight matrices and the control function vary with application. The term $tr\left(\norm{M_{\Sigma}}^2_{\Sigma_k}\right) = tr\left(M_{\Sigma}^T\Sigma_kM_{\Sigma}\right)$, returns the marginal covariance of the robot location. Therefore, $M_{\Sigma} = \tau \bar{M}_{\Sigma}$, where $\tau$ is a positive scalar and $\bar{M}_{\Sigma}$ is a matrix filled with zero or identity entries.
$P(\mathcal{C})$ represents the probability of collision and $M_C$ penalizes the belief states with higher collision probabilities. 

The failure to find an $\epsilon-$safe configuration might be because such a configuration indeed does not exist or simply because there was not enough samples in the roadmap. In such scenario the roadmap has to be extended. Different strategies could be implemented to efficiently extend the roadmap but is not the main focus of the current paper. Therefore we follow a straightforward approach to add more samples when an $\epsilon-$safe configuration cannot be found. Given a node from which no $\epsilon-$safe configuration can be found, a circle of certain radius (half the maximum distance allowed between two edges) is drawn. Samples are then added to the roadmap and the PRM graph is updated until an $\epsilon-$safe configuration is found or until time-out.

\section{Simulation Results}
\label{results}
In this section we first provide a comparison of our approach with~\cite{park2018IEEE} and~\cite{dutoit2011IEEE}. We then explore the capabilities of our approach in two simulation domains. Performance are evaluated on an Intel{\small\textregistered} Core i7-6500U CPU$@$2.50GHz$\times$4 with 8GB RAM under Ubuntu 16.04 LTS.

\subsection{Comparison to Other Approaches} 
\begin{figure}[]
  \subfloat[]{\includegraphics[scale=0.2]{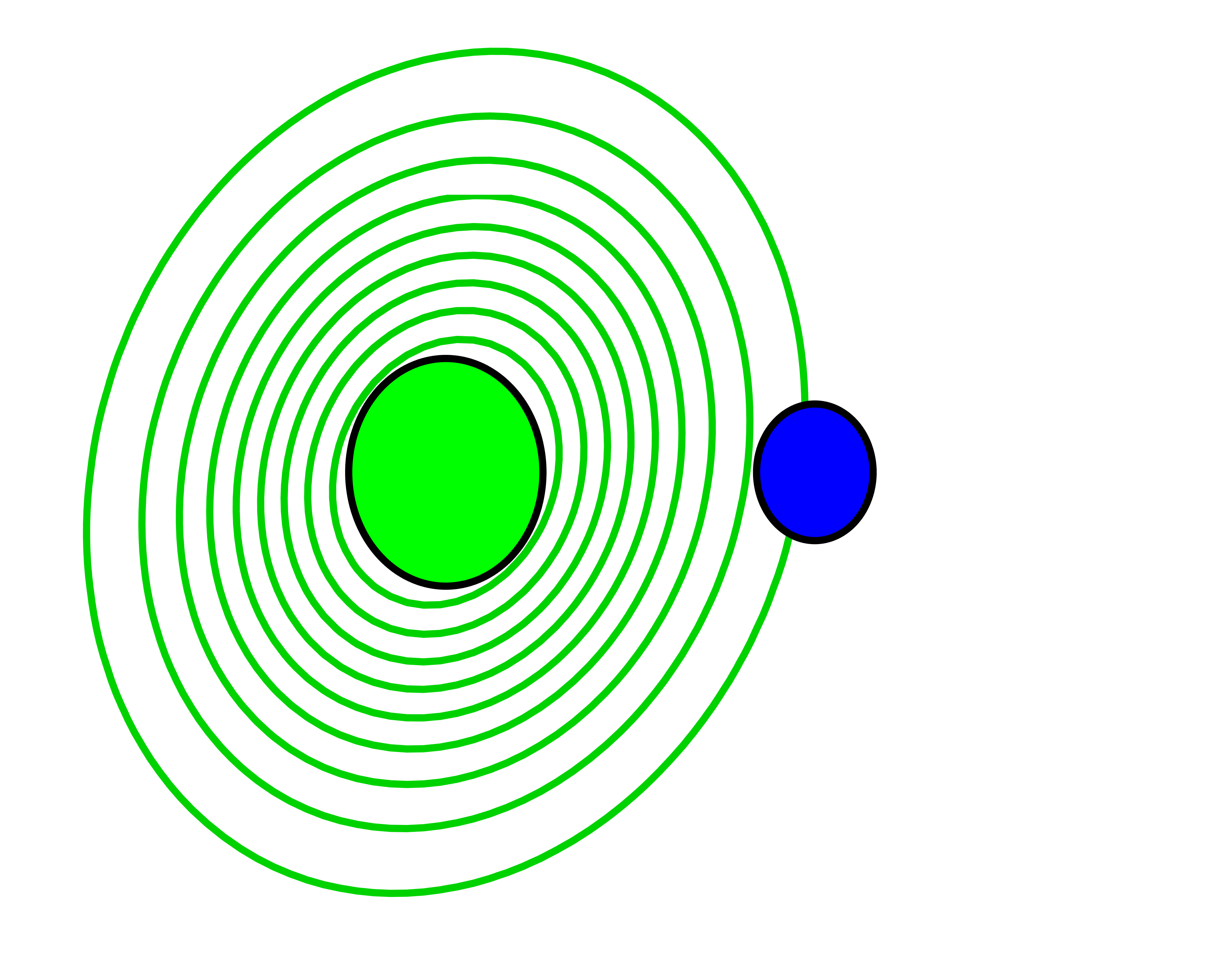}} 
  \subfloat[]{\includegraphics[scale=0.2]{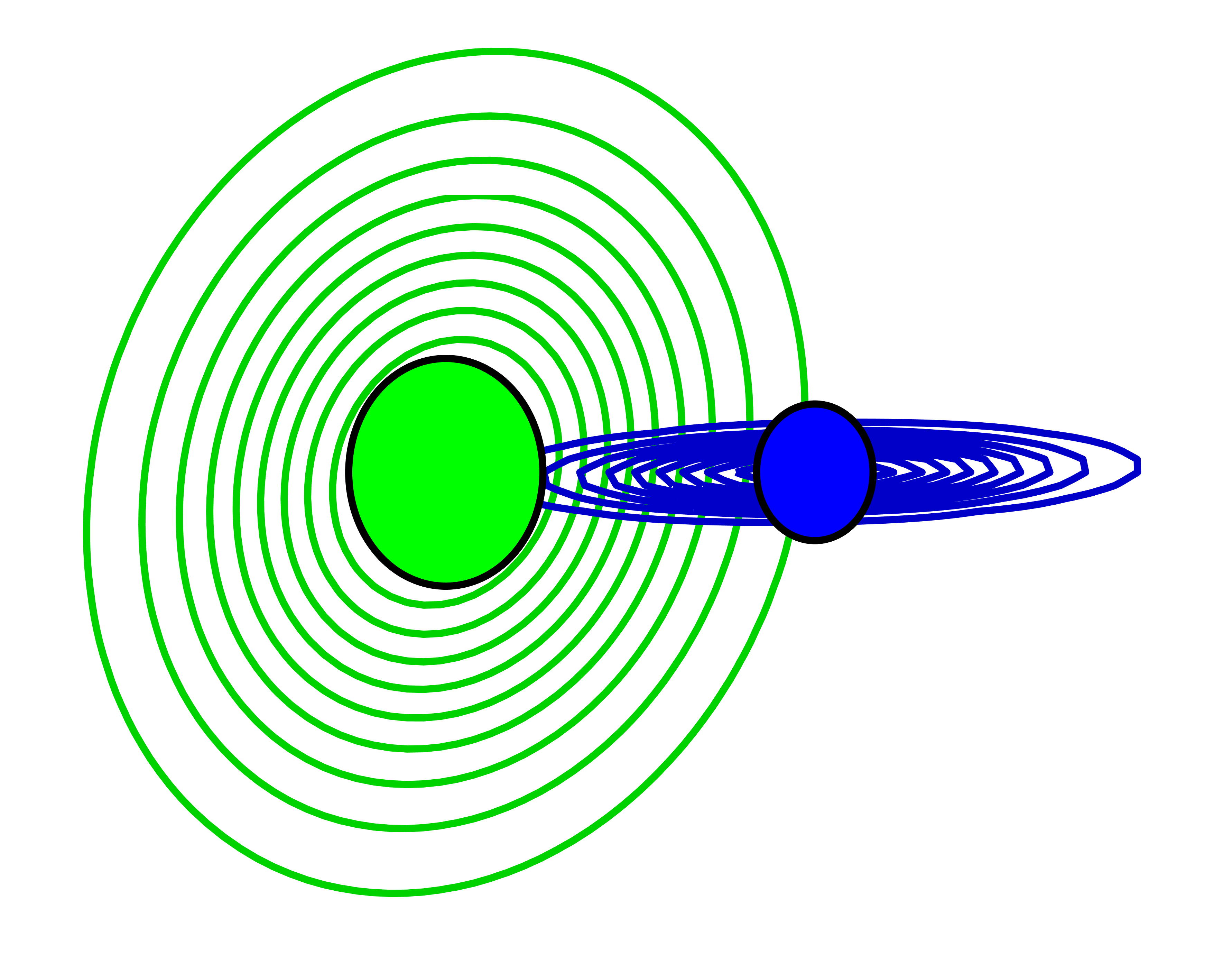}} 
   \subfloat[]{\includegraphics[scale=0.83,height=1.5cm]{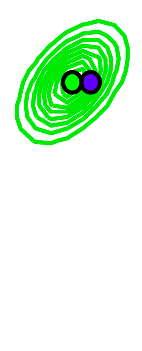}}
  \caption{Comparison of our approach to other methods. (a) The robot state is known perfectly, however the obstacle location is uncertain. (b) Robot state uncertainty is considered (contours in blue). The collision probability value computed with~\cite{park2018IEEE} gave a much higher value. (c) A point-like robot and obstacle are considered. The values computed with~\cite{park2018IEEE,dutoit2011IEEE} are much lower than expected.}
  \label{fig:comparison}
\end{figure}

\begin{table*}[t]
\small\sf\centering
\begin{tabular}{ |c|c|c|c|c| }
\hline
Case & Algorithm & Collision probability & Computation time (s)& Feasible \\ 
 \hline
\multirow{4}{*}{(a)} & Numerical integral & $4.62\%$ & 0.8896 $\pm$ 0.0356 &Yes \\
   & Du Toit and Burdick~\cite{dutoit2011IEEE} & $5.84\%$ &0.0026 $\pm$ 0.0003 &Yes\\ 
   & Park \textit{et al.}~\cite{park2018IEEE} & $33.26\%$ &0.2367 $\pm$ 0.2081 &No \\  
    & ~Our approach & $4.61\%$ & 0.0232 $\pm$ 0.0024 &Yes \\ \hline
 \multirow{4}{*}{(b)} & Numerical integral & $8.25\%$ & 1.2309 $\pm$ 0.0298 &Yes \\ 
   & Du Toit and Burdick~\cite{dutoit2011IEEE} & $14.20\%$ &0.0021$\pm$ 0.0001 & No\\ 
 & Park \textit{et al.}~\cite{park2018IEEE} & $36.31\%$ & 0.2108 $\pm$ 0.3067& No \\
  & ~Our approach & $8.22\%$ &0.0208 $\pm$ 0.0021 &Yes \\ \hline
 \multirow{4}{*}{(c)} & Numerical integral & $14.82\%$ & 1.2450 $\pm$ 0.0301 &No  \\
 & Du Toit and Burdick~\cite{dutoit2011IEEE} & $0.46\%$ & 0.0019 $\pm$ 0.0004 &Yes\\ 
 & Park \textit{et al.}~\cite{park2018IEEE} & $0.61\%$ & 0.3145 $\pm$ 0.4610&Yes \\
   & ~Our approach & $14.83\%$ & 0.0271 $\pm$ 0.0087 &No \\ \hline
\end{tabular}
\caption{Comparison of collision probability methods.}
\label{tab:comparison}
\end{table*}

Park \textit{et al.}~\cite{park2018IEEE} approximate the integral in (\ref{eq:collision_prob}) as $Vp(\B{x}_k,\B{s})$, where $V$ is the 2D footprint or area occupied by the robot. For computing $p(\B{x}_k,\B{s})$, they first assume a distribution centered around the obstacle with the covariance being the sum of the robot and obstacle location uncertainties. The collision probability is then computed by finding the $\B{x}_k$ that maximizes $p(\B{x}_k,\B{s})$ and formulate the problem as an optimization problem with a Lagrange multiplier. In~\cite{dutoit2011IEEE} the density of the center of the robot is used. For comparing with these approaches, we formulate the problem as given in each of these works\footnote{For the comparison, the approaches in~\cite{dutoit2011IEEE,park2018IEEE} have been reproduced to the best our understanding and the reproduced codes (including numerical integration and our approach) can be found here--- \url{https://bitbucket.org/1729antony/comparison_cp_methods/src/master/}}. In order to validate the values computed using our approach, we perform numerical integration of the expression in (\ref{eq:collision_prob}), which gives the exact collision probability value. 

Three different cases are considered as shown in Fig.~\ref{fig:comparison}. The solid green circle denotes an obstacle of radius 0.5m and its corresponding uncertainty contours are shown as green circles. The solid blue circle denotes a robot of radius 0.3m with the blue circles showing the Gaussian contours. We define a collision probability threshold of $0.1$, that is, a $0.9-$safe configuration. The collision probability values
and the computation times are provided in Table~\ref{tab:comparison}. In Fig.~\ref{fig:comparison}(a), the robot position known with high certainty and our approach computes collision probability as $4.61\%$ and hence the given configuration is a $0.9-$safe configuration. The numerical integral provides the actual value and as seen in Table~\ref{tab:comparison}, it is computed to be $4.62\%$, thus proving the exactness of our method. However, the collision probability computed as given in~\cite{park2018IEEE} is $33.26\%$ (almost seven times our value), predicting the configuration to be unsafe. The approach in~\cite{dutoit2011IEEE} gave the value of $5.84\%$, a much tighter upper bound. In Fig.~\ref{fig:comparison}(b), there is robot uncertainty along the horizontal axis and the collision probability computed using our approach is $8.22\%$. The actual value is computed to be $8.25\%$. As compared to the previous case, the probability has almost doubled. This is quite intuitive as seen from the robot uncertainty spread and hence there is greater chance for intersection between the robot and the obstacle. The value computed using the approach in~\cite{park2018IEEE} is $36.31\%$ ($4.5$ times our value). The approach in~\cite{dutoit2011IEEE} also gave a higher value of $14.20\%$. Unlike the approaches in~\cite{dutoit2011IEEE,park2018IEEE} our approach rightly predicts the configuration to be a $0.9-$safe configuration. The higher values obtained using~\cite{dutoit2011IEEE,park2018IEEE} are due to the overly conservative nature of the estimates. 

The approach of Park \textit{et al.}~\cite{park2018IEEE} and~\cite{dutoit2011IEEE} assumes that the robot radius is very small. We also compute the collision probabilities for a robot and an obstacle with radius $0.05$m each, where the robot and the obstacle are touching each other (Fig.~\ref{fig:comparison}(c)). The obstacle location is also much more certain, with the uncertainty reduced by $97\%$ as compared to cases in Fig.~\ref{fig:comparison}(a),(b). Actual value obtained using numerical integral is $14.82\%$. The probability of collision computed using our approach is $14.83\%$, whereas, using the approach in~\cite{park2018IEEE} the computed value is $0.61\%$ and the approach in~\cite{dutoit2011IEEE} computes it to be $0.46\%$. Thus our approach predicts the configuration to be unsafe. To get a sense of the actual value, we compute the area of the covariance matrix, which is $6.28 \times 10^{-4}m^2$. This clearly indicates that $0.61\%$ is too small a value and the configuration is not $0.9-$safe configuration. {Using the approaches in~\cite{dutoit2011IEEE,park2018IEEE} would lead to collision as it predicts the configuration to be safe. Our approach computes the exact probability of collision and outperforms the approaches in~\cite{dutoit2011IEEE,park2018IEEE}.

\subsection{2D Environment Domain}

\begin{figure*}[t]
\centering
  \subfloat[]{\includegraphics[scale=0.22]{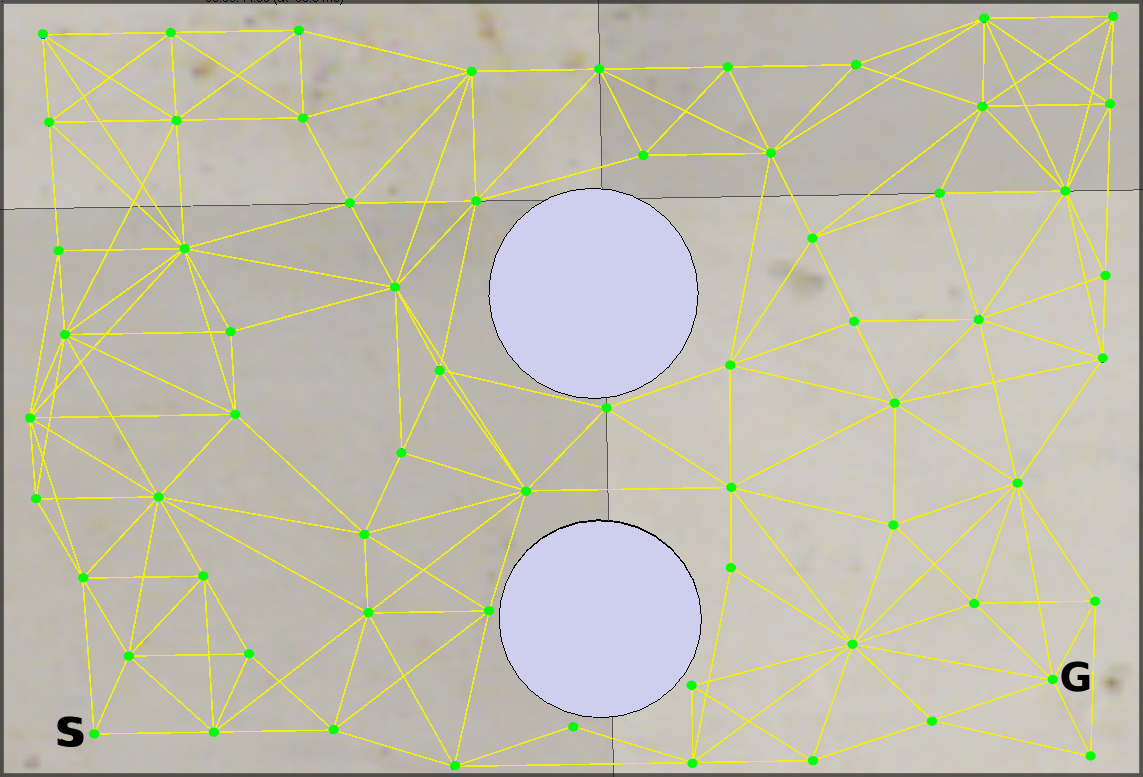}} \hspace{0.3cm}%
  \subfloat[]{\includegraphics[scale=0.25]{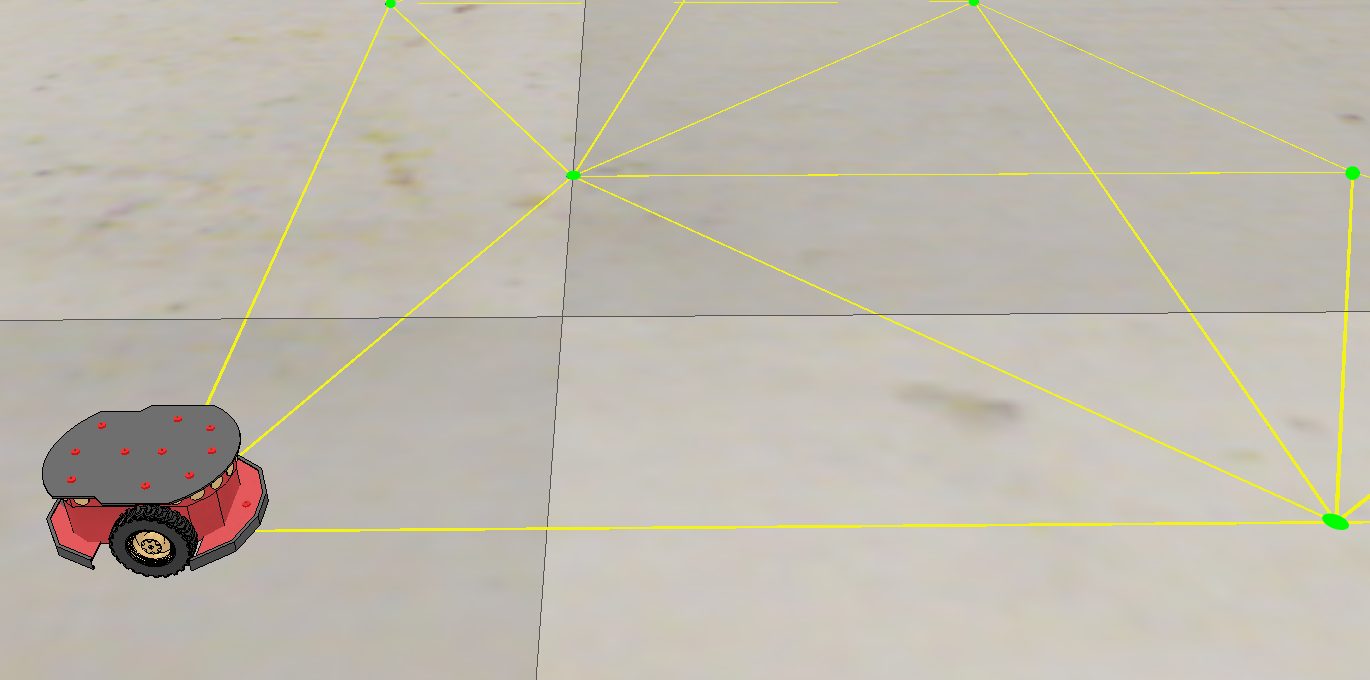}}
  \caption{Simulation environment. (a) Scaled-down ($\times \frac{1}{4}$) top view of the environment with the sampled roadmap and start and goal locations of the robot. (b) Pioneer robot at the starting node of the roadmap. }
  \label{fig:env}
\end{figure*}

\begin{figure}[h!]
\centering
  \subfloat[]{\includegraphics[scale=0.275]{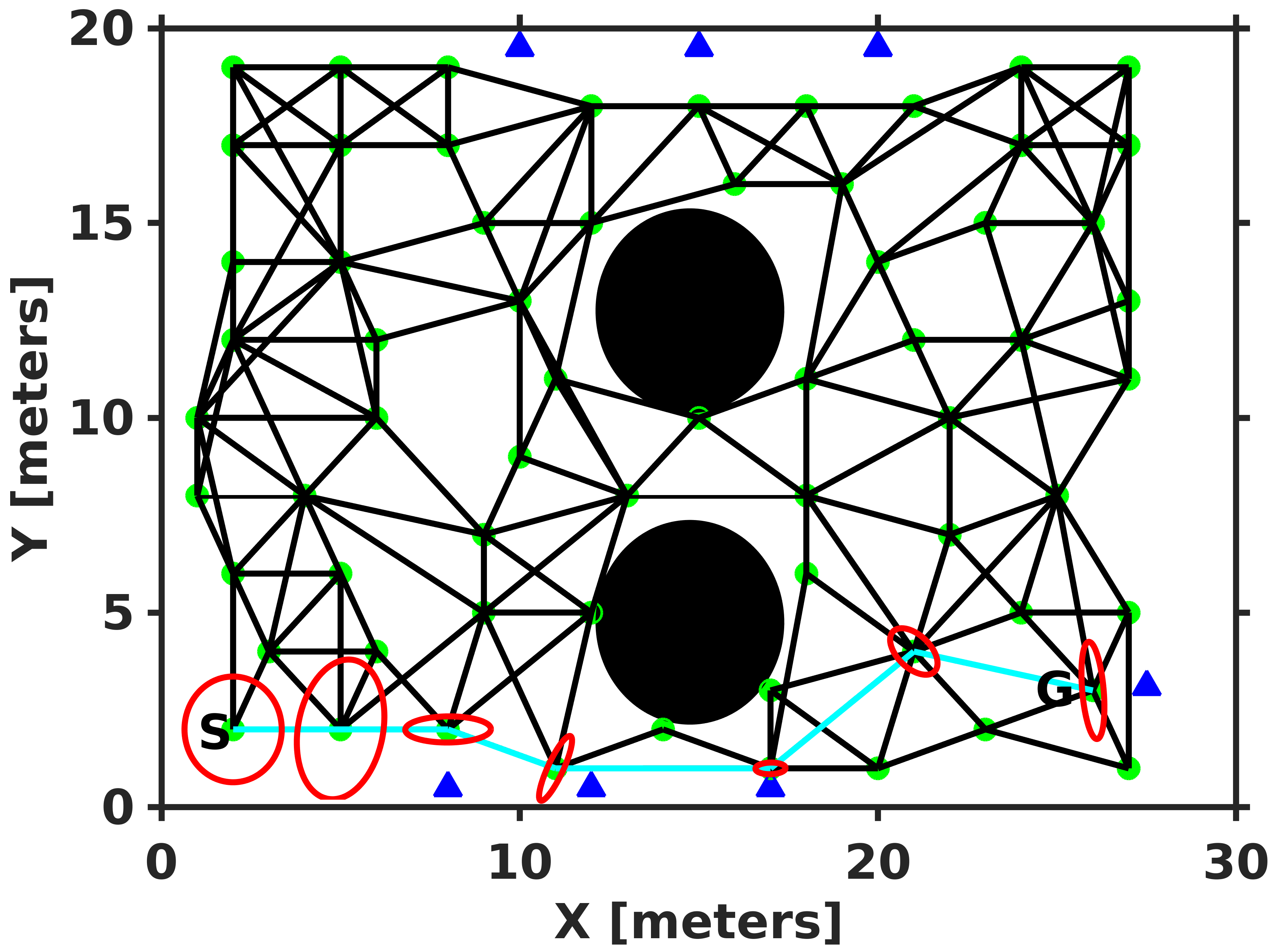}} 
  \subfloat[]{\includegraphics[scale=0.275]{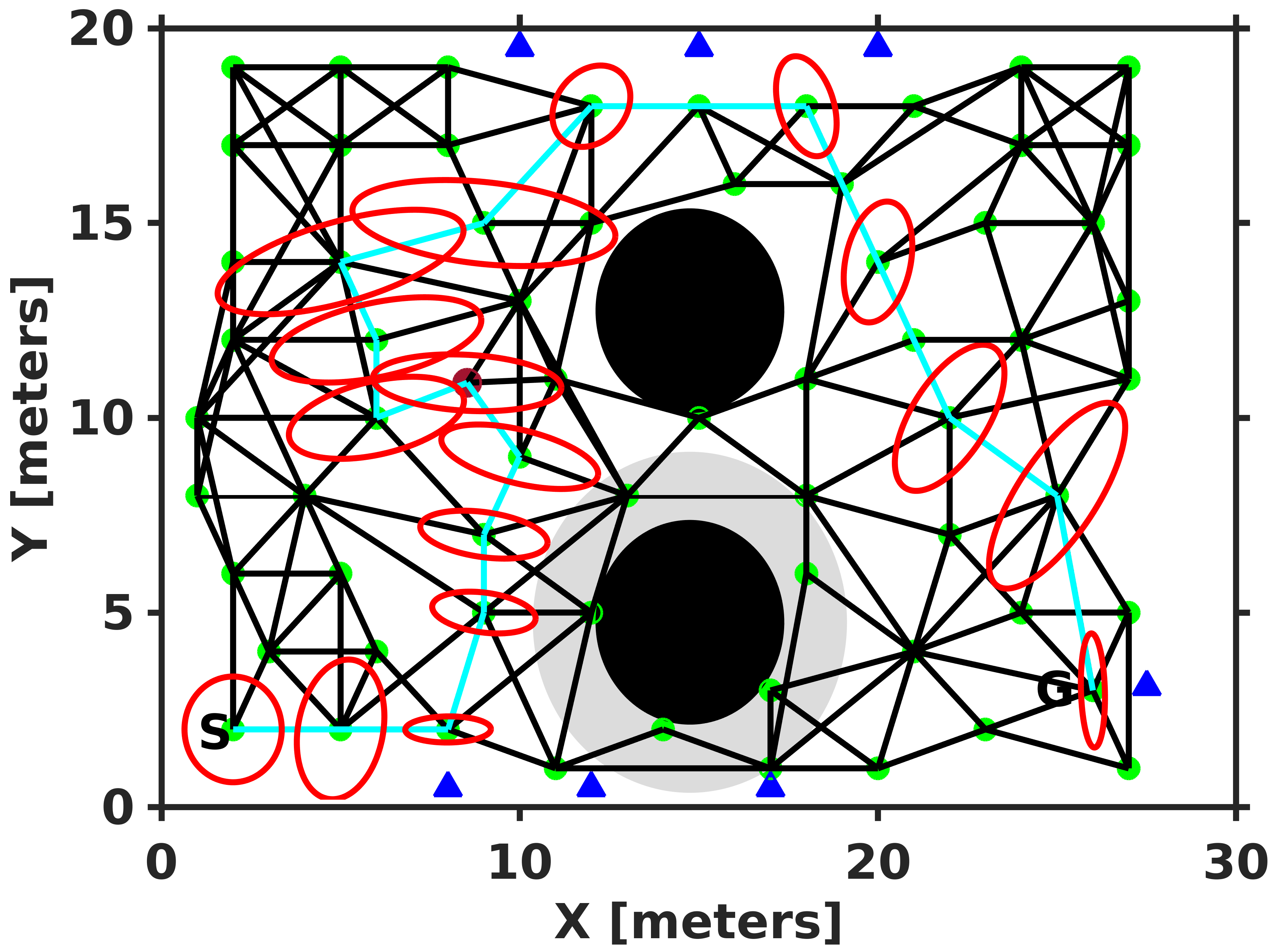}}
  
    \subfloat[]{\includegraphics[scale=0.275]{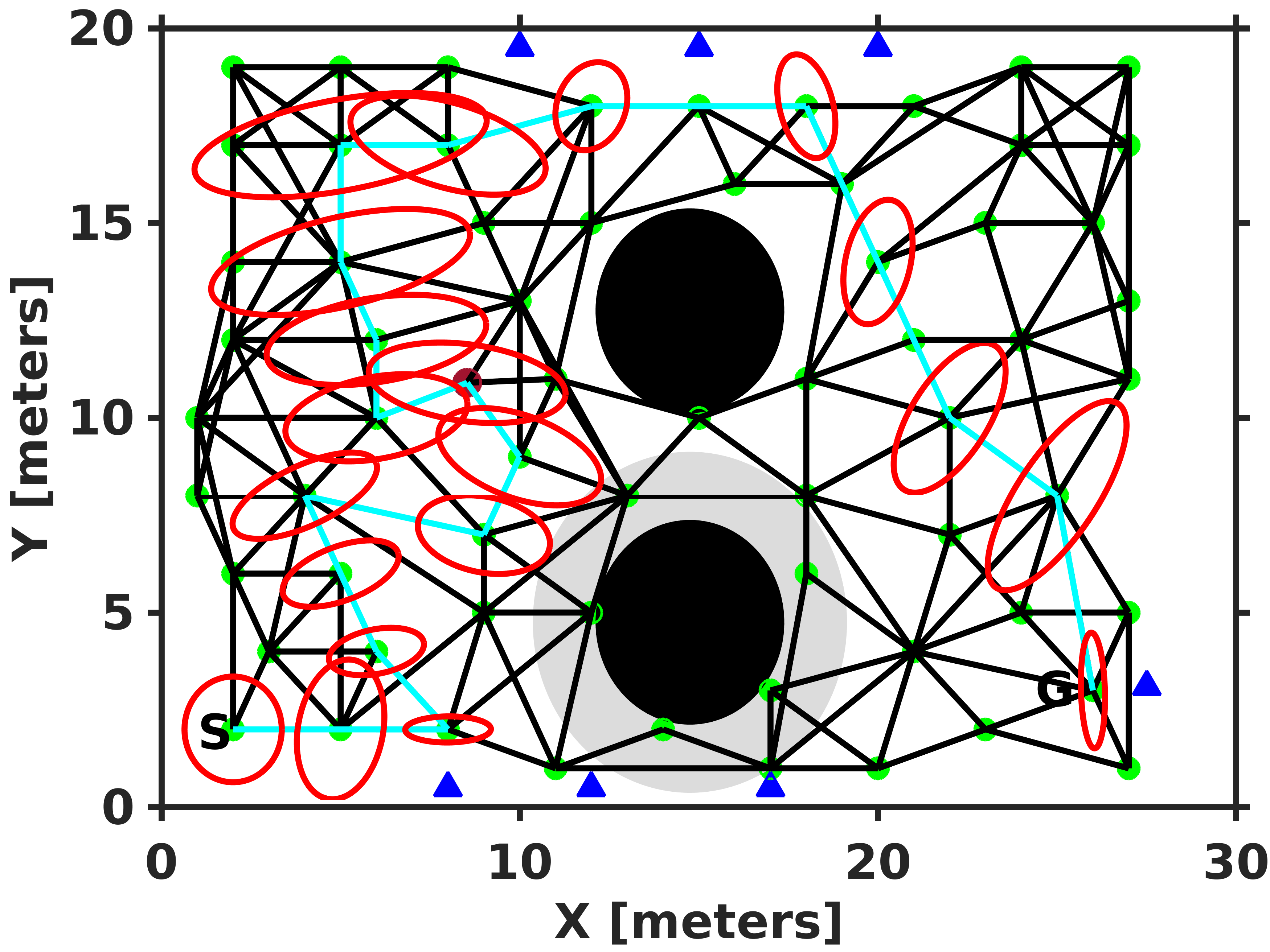}}
    \subfloat[]{\includegraphics[scale=0.275]{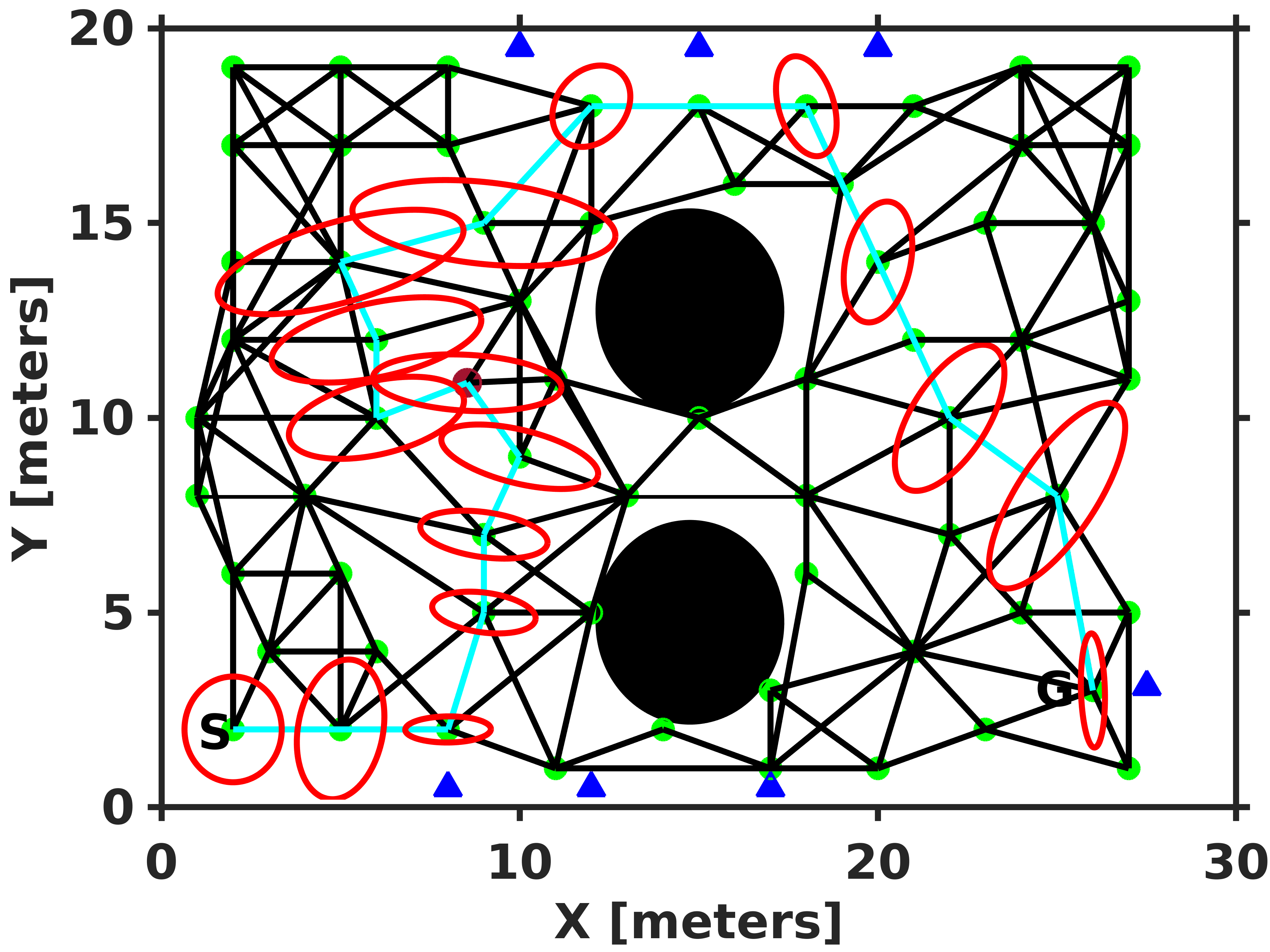}}
    
   \subfloat[]{\includegraphics[scale=0.275]{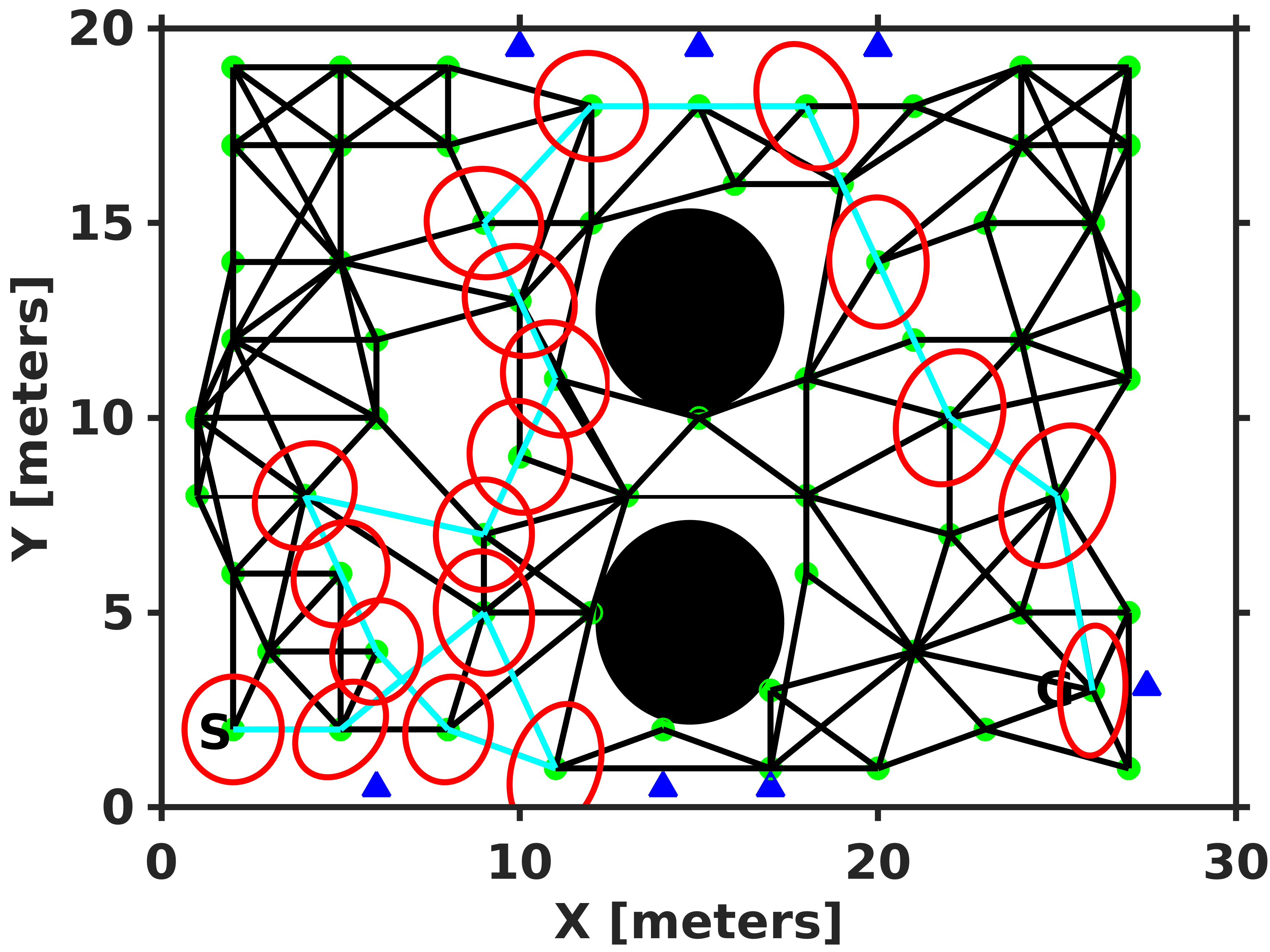}}
      \subfloat[]{\includegraphics[scale=0.275]{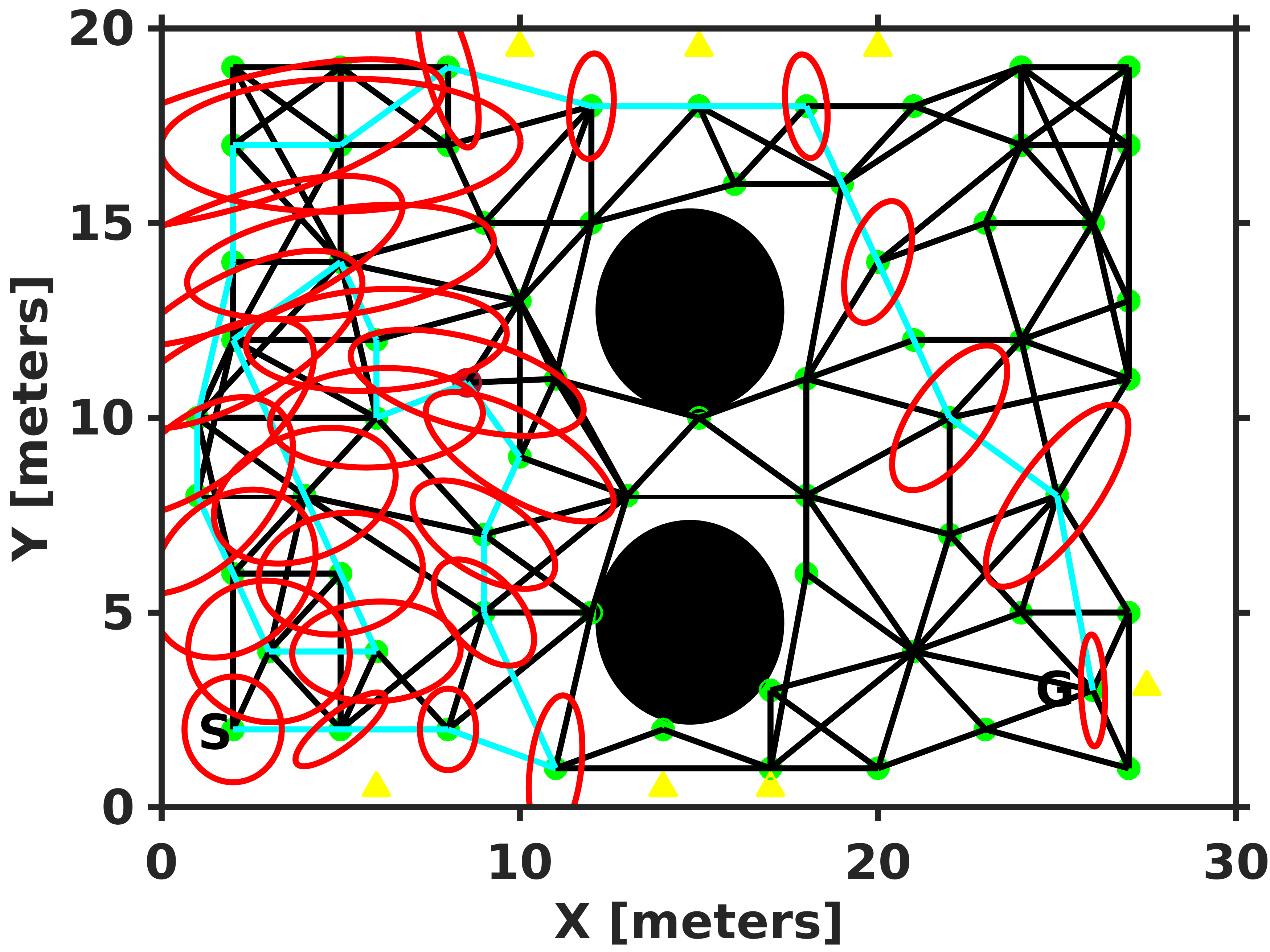}}
  \caption{Trajectory and the covariance evolution for single planning instantiations are shown. Different cases with obstacle uncertainty for a point robot and a robot of radius 0.3$m$ are shown in (a), (b), (c) and (d). (e) The planned trajectory when there is uncertainty in beacon locations. (f) True beacon locations are shown in yellow.}
  \label{fig:collision_2Drobot}
   \end{figure}
   
   \begin{table*}[t]
\small\sf\centering
\begin{tabular}{ |c|c|c|c|c| }
\hline
Approach & Robot radius & Obstacle uncertainty & Beacon (object) uncertainty & Planned trajectory \\ 
 \hline
 Our & Point & No & No & Fig.~\ref{fig:collision_2Drobot}(a)\\
 \hline
 Our & Point & Yes & No & Fig.~\ref{fig:collision_2Drobot}(b)\\
 \hline
 Our & 0.3 m & No & No & Fig.~\ref{fig:collision_2Drobot}(a)\\
 \hline
\cite{dutoit2011IEEE} & 0.3 m & No & No & Fig.~\ref{fig:collision_2Drobot}(a)\\
  \hline
     \cite{park2018IEEE} & 0.3 m & No & No & Fig.~\ref{fig:collision_2Drobot}(d)\\
  \hline
   Our & 0.3 m & Yes & No & Fig.~\ref{fig:collision_2Drobot}(c)\\
   \hline
   Our & 0.3 m & No & Yes & Fig.~\ref{fig:collision_2Drobot}(e)\\
   \hline
   Our & 0.3 m & No & No (true beacon location) & Fig.~\ref{fig:collision_2Drobot}(f)\\
   \hline
   Our & 0.3 m & No & No (mean beacon location) & Fig.~\ref{fig:collision_2Drobot}(a)\\
   \hline
\end{tabular}
\caption{Different configurations used for the 2D environment domain.}
\label{tab:cases}
\end{table*}

We consider the case of a environment where a mobile robot moving in an environment of $30m \times 20m$. A scaled-down top view is seen in Fig.~\ref{fig:env}(a). The underlying PRM graph, the start (S in the figure) and goal (G in the figure) locations can also be seen. The gray circles denote the obstacles in the environment. Fig.~\ref{fig:env}(b) shows a Pioneer P3DX robot at the start location. For the robot motion model, we consider the following non-linear dynamics~\cite{thrun2005book} 
\begin{equation}
\begin{split}
x_{k+1} & = x_{k} + \delta_{trans}  \cos(\theta_{k}+ \delta_{rot1})\\
y_{k+1} & = y_{k} + \delta_{trans}  \sin(\theta_{k}+ \delta_{rot1})\\
\theta_{k+1} & = \theta_{k} + \delta_{rot1}+ \delta_{rot2}
\end{split}
\label{odometry_model}
\end{equation}

\noindent where $\B{x}_k\doteq(x, y, \theta)$ is the robot pose at time $k$ and $\B{u}_k \doteq (\delta_{rot1}, \delta_{trans}, \delta_{rot2})$ is the applied control. The model assumes that the robot ideally implements the following commands in order: rotation by an angle of $\delta_{rot1}$, translation of $\delta_{trans}$ and a final rotation of $\delta_{rot2}$ orienting the robot in the required direction. The robot accrue translational and rotational errors while executing $\B{u}_k$ and localizes itself by estimating its position using signal measurements from beacons $\bar{b}_1, \ldots, \bar{b}_7$, which are located at $(x_{\bar{b}_1},y_{\bar{b}_1}), \ldots, (x_{\bar{b}_7},y_{\bar{b}_7})$. The signal  strength decays quadratically  with the  distance to the beacon, giving the following observation model with sensor noise $v_k$,
\begin{equation}
\B{z}_k =
\begin{bmatrix}
 1/\left((x_k - x_{\bar{b}_1})^2 + (y_k - y_{\bar{b}_1})^2 + 1\right) \\
 \vdots  \\
 1/\left((x_k - x_{\bar{b}_7})^2 + (y_k - y_{\bar{b}_7})^2 + 1\right)
\end{bmatrix}
+ v_k 
\end{equation}

We validate our approach in the above discussed environment by varying different parameters, a summary of which is provided in Table~\ref{tab:cases}. Below we detail each of cases considered in Table~\ref{tab:cases}. We first consider the motion planning approach for a point-like robot. The cost function is of the form in~(\ref{eq:cost}) with $M_{u} = 0.3$, $M_g = diag(0.8, 0.8)$, $M_{\Sigma} = diag(1, 1)$ and $M_C = 10$. The underlying PRM graph with 65 nodes is shown in Fig.~\ref{fig:collision_2Drobot}, with the green dots denoting the sampled nodes. The robot, starting from its initial belief state (mean pose denoted by S in the figure) has to reach the node $\B{x}_g$ (G in the figure), while reducing its uncertainty. The blue triangles denote the beacons that aid in localization. The solid black circles with radius 0.5m, represent obstacles in the environment and the red ellipses denote the 3$\sigma$ covariances (only the ($x$,$y$) portion is shown). Unless otherwise mentioned, in all the experiments, $0.99-$safe configurations are solicited and the total planning time is the average time for 25 different runs.

We first consider a case with a point robot and no uncertainty in obstacle location. The planned trajectory in this case is seen in cyan in Fig.~\ref{fig:collision_2Drobot}(a) with total planning time of $0.0051s(\pm0.0008s)$. Please note that the total planning time also includes the collision probability computation time. Next, we consider uncertainty in one of the obstacle location, whose covariance ellipse is shown in gray. The planned trajectory is seen in cyan in Fig.~\ref{fig:collision_2Drobot}(b) and the planning was completed under $0.0279s(\pm0.0043s)$. Due to the uncertainty in the obstacle location, the robot takes a longer route to avoid collision. A robot of radius 0.3$m$ and certain (negligible uncertainty) obstacles gave the same trajectory as in Fig.~\ref{fig:collision_2Drobot}(a) with a planning time of $0.0055s(\pm0.0009s)$. However, when the obstacle location is uncertain the resulting trajectory is as shown in Fig.~\ref{fig:collision_2Drobot}(c). A change in the trajectory is observed, as compared to the case of a point robot in Fig.~\ref{fig:collision_2Drobot}(b). The planning time in this case is $0.0294s(\pm0.0047s)$. It is also worth mentioning that in Fig.~\ref{fig:collision_2Drobot}(b) and (c), the roadmap was updated by adding a node since a $0.99-$safe configuration could not be found. The added node is seen in brown, with its coordinates being approximately $(9,11)$. We also run the case with no obstacle uncertainty and a robot of radius 0.3$m$ using the approach of Park \textit{et al.}~\cite{park2018IEEE}. In this case the planned trajectory is as given in Fig.~\ref{fig:collision_2Drobot}(d). Note that using our approach, the same scenario gives a shorter trajectory (Fig.~\ref{fig:collision_2Drobot}(a)). The longer trajectory computed using the approach in~\cite{park2018IEEE} is due to the fact that a loose upper bound is computed for the collision probability. As a result a longer trajectory is obtained. Contrary to this, we compute the exact collision probability and hence a shorter trajectory is synthesized. The same scenario is also run with the approach in~\cite{dutoit2011IEEE} and produced a trajectory similar to ours. However, since the uncertainties are significantly lower, the approximate collision probability values computed using~\cite{dutoit2011IEEE} are much smaller than the actual values.

Next, we consider the case with uncertainty in the location of the beacons. The considered robot radius is 0.3$m$ with the bottom obstacle being uncertain with covariance $diag(0.49,0.49)$. Taking object uncertainty into account, the planned trajectory with covariance evolution is as shown in Fig~.\ref{fig:collision_2Drobot}(e). Fig.~\ref{fig:collision_2Drobot}(f), shows the trajectory planned with true beacon locations. The beacons are shown in yellow to denote the true location. Considering only the mean position of the beacons and neglecting the position uncertainty, the planned trajectory is as shown in Fig.~\ref{fig:collision_2Drobot}(a). Actual execution of this would lead to collision with the bottom obstacle. However, executing the planned trajectory obtained by considering the uncertainty in beacon locations does not violate the $\epsilon-$safety criterion and all the configurations are $0.99-$safe. 

It is noteworthy that though we have discussed a 2D environment, the approach directly extends to  a mobile robot navigating in a 3D environment. In such domains, the mobile robot may be represented by a minimum volume enclosing sphere. Similarly, the obstacles can also be approximated by their corresponding minimum volume enclosing spheres. Hence the collision condition is the same as given in~(\ref{eq:coll_condition}) and therefore the approach discussed in this paper remains valid. 



\subsection{Laser-grasp Domain}

\begin{figure}[t]
\centering
  \subfloat[]{\includegraphics[scale=0.26]{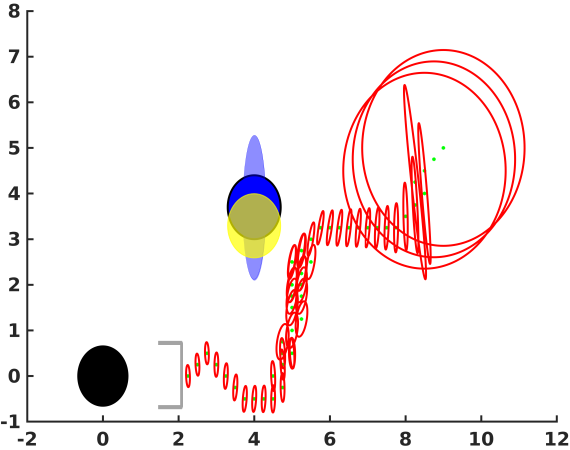}} \hspace{0.2cm}%
  \subfloat[]{\includegraphics[scale=0.26]{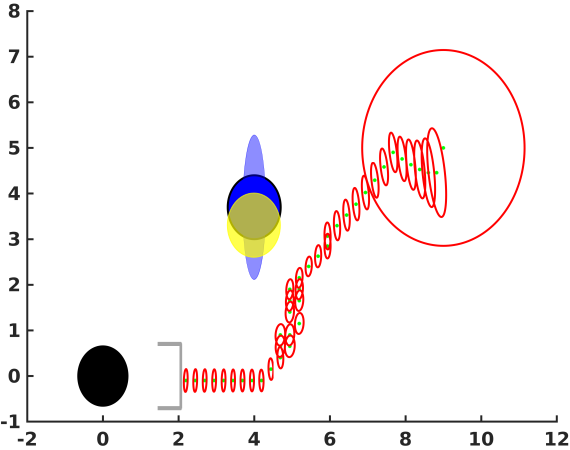}}
  \caption{Trajectory of the end-effector; green dots denote its mean and the red ellipses denote the covariance matrix. The puck is shown in black and the end-effector is shown to its right. (a) Trajectory and covariance evolution when object uncertainty is not considered and (b) when object uncertainty is considered. }
  \label{fig:with_and_without}
\end{figure}

\begin{figure}[t]
\centering
  \subfloat[]{\includegraphics[scale=0.28]{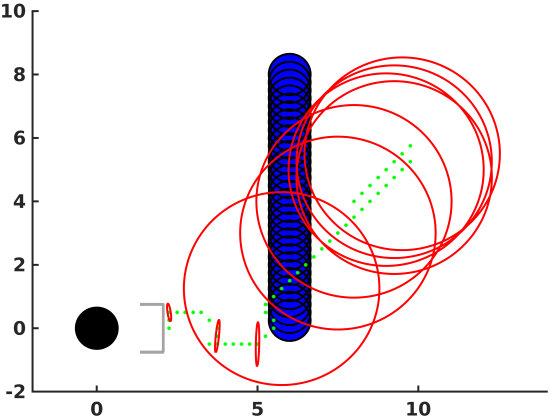}} \hspace{0.2cm}%
  \subfloat[]{\includegraphics[scale=0.265]{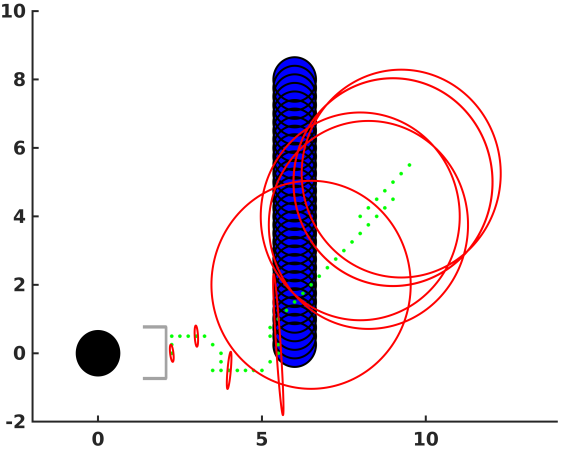}}
  \caption{Green dots denote the mean of the state trajectory and the red ellipses denote the covariance matrix. Mean position of the obstacle at each time instant is visualized in blue. (a) State trajectory and covariance evolution during offline collision avoidance planning. (b) More information is acquired during online planning, reducing the uncertainty of the obstacle and thereby leading to a change in the planned trajectory.}
  \label{fig:collision}
\end{figure}

We consider two modified versions of the laser-grasp domain as suggested in~\cite{platt2010RSS}. In this domain, a planar robot manipulator must locate and proceed towards a round puck. The state space is the position of the manipulator's end-effector relative to a grasping point defined directly in front of the puck. Though the end-effector position is assumed to be known completely, the state is not directly observed since the puck position is unknown. Its position can be determined using the laser range finder that points out as a horizontal line from the end-effector. The underlying system dynamics is
 \begin{equation}
f(\B{x}_t,\B{u}_t) = \B{x}_t \ + \ \B{u}_t
\end{equation}  

\noindent where $\B{x} \in \mathbb{R}^2 $ denotes the state space and $\B{u} \in \mathbb{R}^2$ is the end-effector velocity. The cost function is of the form in~(\ref{eq:cost}) with $M_{u} = diag(10,10)$, $M_g = diag(100, 100)$, $M_{\Sigma} = diag(10000, 10000)$ and $M_C = 10$.

First, we consider a scenario wherein an additional object is placed that aids in localization. In this scenario, the state is the end-effector position which is not known precisely due to actuation errors. The goal is to place the end-effector directly in front of the puck so as to be able to grasp it. Both the object and the puck can be detected by the horizontal laser. However, the object location is not known exactly and the $3\sigma$ uncertainty ellipse is shown in light blue in Fig.~\ref{fig:with_and_without}(a) and (b). The mean position is visualized by the blue blob and the yellow blob denotes the actual object location. The red ellipses represent the state covariance at different points along the trajectory. Fig.~\ref{fig:with_and_without}(a) shows the case in which object uncertainty is not considered and the object is assumed to be its mean position. The manipulator moves towards the object first, localizing the end-effector position and then proceeds further to place the end-effector at the grasping point. However, as seen in Fig.~\ref{fig:with_and_without}(a), while executing this plan produced offline, not considering the object uncertainty leads to the collision of the end-effector with the true object (in yellow). When the object uncertainty is considered, the execution of the plan do not lead to collision, as it can be seen in Fig.~\ref{fig:with_and_without}(b). This illustrates the fact that not considering object uncertainty can wrongly localize the robot, leading to catastrophes. 

Next, we consider a scenario wherein the state space is the position of the manipulator's end-effector relative to a grasping point defined directly in front of the puck. The state is not directly observed since the puck position is unknown. However, as soon as the manipulator starts to move, a ball starts to roll in between the manipulator and the puck. The ball follows a Gaussian velocity distribution, and therefore at each time instant, the mean position of the ball and the corresponding uncertainty can be estimated. The mean position of the ball at each time instant is shown in blue in Fig.~\ref{fig:collision}(a) and (b). The green dots denote the mean of the state trajectory. As seen in  Fig.~\ref{fig:collision}(a), the manipulator initially moves downwards. However, as the ball comes closer, the manipulator retraces its path and move upwards towards its starting position to avoid collision. This is so because the safety constraint for $\epsilon = 0.99$ is violated. As the ball keeps moving upwards, after a while, it is seen that the manipulator takes a downward action just before reaching its starting position since the configuration is a $0.99-$safe configuration.

The scenario in Fig.~\ref{fig:collision}(b) is similar to that of Fig.~\ref{fig:collision}(a). However, it is seen that once the manipulator retraces its path backward towards the starting position, it takes a downward action much earlier. This is because more information is acquired during online planning and the uncertainty bound on the obstacle changes with time. 

The 2D manipulator domain studied here directly extends to 3D manipulator scenarios for both static and mobile manipulators. In the case of static manipulators, the end-effector is approximated as a sphere. Each link is approximated as a set of spheres kept side by side. However, in heavily cluttered environments such an approximation can be computationally intensive since each sphere has to be checked for collision with obstacles. An alternative and effective approach is to consider the minimum-volume enclosing ellipsoid for each link~\cite{rimon1997JINT}. It is known that for every convex polyhedron, there exits a unique ellipsoid of minimal volume that contains the polyhedron and is called the \textit{L\"{o}wner-John ellipsoid} of the polyhedron~\cite{grotschel1988geometric}. Thus each link can be represented by their corresponding L\"{o}wner-John ellipsoids. The distance between two ellipsoids is used to modify the collision condition in~(\ref{eq:coll_condition}). For mobile manipulators, the collision condition should also checked for the base as discussed in the 2D robot section.

\section{Conclusion}
In this paper, we have addressed a novel approach to compute the probability of collision under robot and obstacle pose uncertainties. The collision probability is computed as an infinite series whose convergence is proved. An upper bound for the truncation error is also derived. As shown in Fig.~\ref{fig:convergence}, convergence analysis is performed for different configurations and it is seen that our approach is of the order of milliseconds and therefore can be used in online planning. We also provide a comparison with the approaches in~\cite{park2018IEEE,dutoit2011IEEE}. In addition, we incorporate landmark uncertainties in belief space planning and derive the resulting Bayes filter in terms of the prediction and measurement updates of the EKF. Finally, experimental evaluation for a mobile robot scenario and a 2D manipulator is performed to illustrate our approach. We have considered static obstacles in this paper and the immediate future work is to realize the approach in simulated and real-world environments with dynamic obstacles.


%
%


\bibliographystyle{plain}
\bibliography{References}

\end{document}